\documentclass[lettersize,journal]{IEEEtran}
\usepackage{amsmath,amsfonts}
\usepackage{algorithmic}
\usepackage{algorithm}
\usepackage{array}
\usepackage[caption=false,font=normalsize,labelfont=sf,textfont=sf]{subfig}
\usepackage{textcomp}
\usepackage{stfloats}
\usepackage{url}
\usepackage{verbatim}
\usepackage{graphicx}
\usepackage{cite}

\usepackage{ulem}
\usepackage{cuted}
\usepackage{amsfonts,amssymb} 
\usepackage{amsmath}
\usepackage{mathtools}
\usepackage{comment}
\usepackage{bbm}
\usepackage{bm}
\usepackage{multicol}
\usepackage{multirow}
\usepackage{threeparttable}
\usepackage[table]{xcolor}
\usepackage{arydshln}
\DeclareMathAlphabet{\mathsfit}{\encodingdefault}{\sfdefault}{m}{sl}
\SetMathAlphabet{\mathsfit}{bold}{\encodingdefault}{\sfdefault}{bx}{n}
\newcommand{\tens}[1]{\bm{\mathsfit{#1}}}
\usepackage{color}

\usepackage[colorlinks,
            linkcolor=green,       
            anchorcolor=green,  
            citecolor=green,        
            ]{hyperref}

\begin{document}

\title{Dynamic Dense Graph Convolutional Network for Skeleton-based Human Motion Prediction}

\author{Xinshun Wang, Wanying Zhang, Can Wang, Yuan Gao, Mengyuan Liu$^\dagger$

\thanks{Xinshun Wang is with the National Key Laboratory of General Artificial Intelligence, Peking University, Shenzhen Graduate School. Wanying Zhang is with the School of Intelligent Systems Engineering, Sun Yat-sen University. Can Wang is with the Multimedia Information Processing Laboratory at the Department of Computer Science, Kiel University, Germany, the Advanced Institute of Information Technology (AIIT), Peking University, and also with Hangzhou Linxrobot Co., Ltd. Yuan Gao is with the Unit of Computing Sciences (CS), Faculty of Information Technology and Communication Sciences (ITC), Tampere University, Finland. Mengyuan Liu, the corresponding author (e-mail: nkliuyifang@gmail.com), is with the National Key Laboratory of General Artificial Intelligence, Peking University, Shenzhen Graduate School.}}

\markboth{Submission to IEEE Transactions on Image Processing}%
{Shell \MakeLowercase{\textit{et al.}}: A Sample Article Using IEEEtran.cls for IEEE Journals}


\maketitle

\begin{abstract}
Graph Convolutional Networks (GCN) which typically follows a neural message passing framework to model dependencies among skeletal joints has achieved high success in skeleton-based human motion prediction task.
Nevertheless, how to construct a graph from a skeleton sequence and how to perform message passing on the graph are still open problems, which severely affect the performance of GCN.
To solve both problems, this paper presents a Dynamic Dense Graph Convolutional Network (DD-GCN), which constructs a dense graph and implements an integrated dynamic message passing.
More specifically, we construct a dense graph with 4D adjacency modeling as a comprehensive representation of motion sequence at different levels of abstraction. Based on the dense graph, we propose a dynamic message passing framework that learns dynamically from data to generate distinctive messages reflecting sample-specific relevance among nodes in the graph. Extensive experiments on benchmark Human 3.6M and CMU Mocap datasets verify the effectiveness of our DD-GCN which obviously outperforms state-of-the-art GCN-based methods, especially when using long-term and our proposed extremely long-term protocol.
\end{abstract}


\section{Introduction}
\IEEEPARstart{H}{uman} motion prediction has received much interest due to its relevance for many real-world applications such as human robot interaction and autonomous driving and for related research topics such as human action recognition \cite{meng2019sample, ke2018learning,liu2017enhanced,liu2023novel,liu2023temporal}. Skeleton-based human motion prediction aims specifically at making predictions using human skeleton data. With the prevalence of off-the-shelf depth sensors \cite{wang2019a}, skeleton data are becoming easy to collect and quite robust against environmental nuisances such as scene illumination. By representing human motion as a sequence of skeleton poses, predictive models can be built to gain insight into the underlying patterns that characterize the motion \cite{luo2021trajectory,foka2010probabilistic,butepage2018anticipating,rudenko2020human,gulzar2021survey,gui2018adversarial,guo2019human}.

It remains a core problem to better infer future human skeletons. To handle this problem, most early work employs Recurrent Neural Networks (RNNs) to address the sequential nature of human motion 
\cite{fragkiadaki2015recurrent,ghosh2017learning,gopalakrishnan2019neural,liu2020human,guo2022fusion,gui2018adversarial,guo2019human} but the difficulty of training them poses a challenge. Researchers have also found early success with Convolutional Neural Networks (CNNs) \cite{butepage2017deep,li2018convolutional,liu2020trajectorycnn,li2019efficient}, which process skeleton data the same way as data with an underlying grid-like structure such as images, thus overlooking the non-Euclidean properties of human bodies.

Regarding promising approaches for human motion prediction, they include MLP-Mixer and Graph Convolutional Networks (GCN). MLP-Mixer is recently leveraged into human motion prediction \cite{guo2023back} and achieves superior performance while being notably efficient. GCNs adopt a more comprehensive view of human skeletons as graphs. Graphs have the intrinsic capability to model articulated structural relations with nodes and edges. This is especially the case for human skeleton data since we already have prior knowledge regarding the natural connectivity of the joint-bone structure.
Current state-of-the-art human motion prediction methods \cite{mao2019learning,dang2021msr,ma2022progressively} typically apply a GCN to extract features from skeletal joints. Regardless of the instantiation, they follow a general neural message passing framework, which involves the aggregation and update of messages in the graph \cite{gilmer2017neural}. Performances of these GCN-based methods severely rely on two main aspects, namely, graph construction and message passing.

\begin{figure*}[h]
\centering
\includegraphics[width=0.999\textwidth]{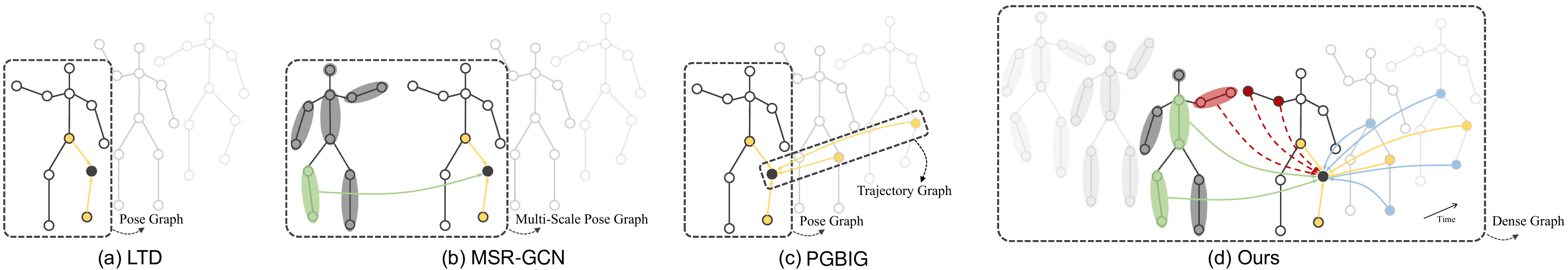}
\vspace{-2em}
\caption{\textbf{Comparison between our method with LTD \cite{mao2019learning}, MSR-GCN \cite{dang2021msr} and PGBIG \cite{ma2022progressively}}. (a) LTD uses spatial message passing on a skeleton graph. (b) MSR-GCN uses spatial message passing on a multi-scale skeleton graph. (c) PGBIG uses separate message passing on a skeleton graph and a trajectory graph. (d) Ours uses integrated dynamic message passing on dense graph. For clarity, edges that connect nodes across different frames are omitted, lines and nodes are assigned different colors corresponding to messages aggregated from different directions, and dotted lines correspond to edges dynamically inferred from input data.}
\label{Figure1}
\vspace{-1em}
\end{figure*}

\textbf{Graph Construction:} GCN-based human motion prediction methods either build a (multi-scale) skeleton graph for spatial modeling or build a trajectory graph for temporal modeling. Specifically, in Learning Trajectory Dependencies (LTD) \cite{mao2019learning}, an implicit skeleton graph (in Fig. \ref{Figure1} (a)) is used to capture spatial dependencies among joints inside of each skeleton. In Multi-Scale Residual (MSR-GCN) \cite{dang2021msr}, the skeleton graph is extended to a multi-scale version (in Fig. \ref{Figure1} (b)) to learn spatial dependencies across multiple scales. In Better Initial Guesses (PGBIG) \cite{ma2022progressively}, an additional trajectory graph (in Fig. \ref{Figure1} (c)) is used to learn temporal dependencies across different skeletons. To encode spatial-temporal dependencies among skeletons, the first question arises:$~$\uline{Instead of using either a spatial graph or a temporal graph, whether a spatial-temporal graph benefits modeling spatial-temporal dependencies}? 

We answer this question by presenting a global perspective for graph construction. Specifically, we construct a dense graph with 4D adjacency modeling (in Fig. \ref{Figure1} (d)) which involves vertices and edges crossing multi-scale spatial-temporal skeletons. Previous (multi-scale) skeleton graphs or trajectory graphs can be treated as special cases of our dense graph. Therefore, our dense graph is expected to be more flexible to model the multi-scale spatial-temporal dependencies.

\textbf{Message Passing:} After the construction of the graph, GCN-based human motion prediction methods use either spatial message passing or separate message passing for feature propagation. Specifically, LTD \cite{mao2019learning} and MSR-GCN \cite{dang2021msr} use spatial message passing on the (multi-scale) skeleton graph. Meanwhile, PGBIG \cite{ma2022progressively} uses separate messages passing on the skeleton graph and the trajectory graph. Considering that these message passing methods are designed for specific graphs, the second question arises:$~$\uline{How to design the specific message passing method for our newly proposed dense graph}? 

We answer this question by designing an integrated dynamic message passing method (in Fig. \ref{Figure1} (d)), which differs from previous message passing methods in two aspects. First, it enables direct message passing across multi-scale spatial-temporal skeletal joints. While previous methods ignore the message passing between the spatial and the temporal skeletal joints.
Second, it enables data-dependant message passing for each skeleton sequence, which further extends the flexibility of our method for spatial-temporal modeling. In contrast, previous message passing methods work in a static manner that shares the parameterized graph among all samples.
Taking both graph construction and message passing into consideration, we develop a Dynamic Dense Graph Convolutional Network (DD-GCN) for skeleton-based human motion prediction. Our DD-GCN consists of multiple Dynamic Dense Graph Convolution (DD-GC) blocks, each of which deploys a multi-pathway design to perform Single-Level Message Passing (SLMP) and Cross-Level Message Passing (CLMP) over our proposed dense graph with 4D adjacency modeling.

Our contributions are three-fold:
(1) To facilitate skeleton-based human motion prediction, we present a dense graph with 4D adjacency modeling to represent the motion sequence at different levels of abstraction as a whole from a global perspective, which enables the message passing layer to efficiently capture long-range spatio-temporal dependencies.
(2) We propose a novel dynamic message passing framework, wherein the aggregators learn dynamically from data and generate informative messages by exploiting sample-specific relevance among nodes in the graph. They do so in a cost-effective way that provides higher interpretability and representational capacity at the expense of very few extra parameters. The overall dynamic dense graph convolutional network (DD-GCN) leverages the two proposals in conjunction, enabling more effective and efficient feature learning.
(3) Extensive experiments on two challenging benchmark datasets, i.e., H3.6M and CMU Mocap, consistently show that our proposed DD-GCN achieves state-of-the-art performances in both traditional long-term (1000 milliseconds) prediction task and our proposed extremely long-term (2000 milliseconds) prediction task.

\section{Related Work}

\subsection{Neural Message Passing on Graphs}

Message Passing Neural Network (MPNN) is a general framework for deep learning on graphs \cite{scarselli2008graph}. GCNs \cite{kipf2016semi} and many other graph networks such as Graph Attention Network (GAT) \cite{velivckovic2017graph} are its special cases. The framework involves the aggregation and update of node features. Since its inception, different flavors of aggregators have been proposed \cite{hamilton2017inductive}. The mean (sum) aggregator simply takes element-wise weighted average (sum) of the feature vectors in the neighborhood. The average or sum operation can be replaced with an alternative reduction function, such as an element-wise maximum or minimum \cite{qi2017pointnet}. In another work \cite{zaheer2017deep} the sum aggregator is combined with MLPs to increase the theoretical capacity of the graph networks. These simple permutation-invariant operations can be parameterized. For example, the weights in the mean aggregator can be set as trainable parameters, as is typical in GCNs \cite{kipf2016semi,schlichtkrull2018modeling,wu2019simplifying,zhao2019t,gao2018large,tong2020directed}. The aggregators discussed above work in a static manner that shares the parameters among all input sample and does not address the sample-specific relations.
Regarding dynamic aggregation, a popular approach is to implement the aggregation using attention \cite{bahdanau2014neural}, which assigns an importance score to each neighbor to weigh its influence. In GAT, the aggregator applies importance scores as weights to define a weighted sum of the neighbors. In another recent approach \cite{chen2021channel}, originally proposed to address human action recognition, the aggregation is implemented in a refinement way, which involves a shared aggregation applied to all channels and multiple channel-wise aggregations corresponding to different channels.

Graph-structured time series data can be treated as a general graph defined in both space and time domain, commonly referred to as a ``spatial-temporal graph'' \cite{yan2018spatial,guo2019attention,wu2019graph,yu2017spatio,li2019spatio,aoun2014graph,herzig2019spatio}.
To fully utilize it, some recent approaches rely on fully convolutional structures to extract spatial-temporal
features from spatial-temporal graphs \cite{yan2018spatial,guo2019attention,yu2017spatio,liu2020disentangling,shi2021adasgn,li2019actional}. Networks of such structures are usually termed as ``spatial-temporal graph convolutional networks''.
Implementation of spatial-temporal graph convolutions in existing work mainly fall into two types. The first type primarily focus on spatial graph convolution for spatial modeling, with additional temporal convolutions for temporal modeling. \cite{yan2018spatial,li2019actional,shi2021adasgn,shi2019skeleton}. The second type represents a unified approach that captures spatial-temporal correlations directly using graph convolution \cite{liu2020disentangling,gao2019optimized}.

\subsection{Skeleton-Based Human Motion Prediction}

Methods addressing the skeleton-based human motion prediction task need to model the relationships between the past, future events and incomplete observations. 
Earlier approaches are based on representation learning, inspired a line of works from skeleton action analysis \cite{xu2023spatiotemporal,xu2023pyramid,shu2022multi,xu2022x,shu2021spatiotemporal,liu2019feature}.
Derived as a generalization of convolutions to non-Euclidean data \cite{bruna2013spectral}, Graph Convolutional Networks have become the favorite choice of backbone for human motion prediction.
In this subsection, we give a presentation on GCN-based approaches in human motion prediction, with relevant exemplary research papers from two perspectives on the matter.

\paragraph{Graph Construction}
LTD \cite{mao2019learning} represents the motion sequence as an implicit fully-connected graph in trajectory space where each trajectory of a coordinate corresponds to a node with encoded temporal information as node features. While they blazed a trail for the use of GCNs in the area, they do not exploit the kinematic properties of human bodies such as connectivity of joints and kinetic chain of body segments.
LDR \cite{cui2020learning} introduces a pose graph with a predefined topology based on natural adjacency, which is regarded as supplementary to the implicit graph.
Researchers also seek to exploit the hierarchical anatomical relations underlying human motion. To this end, there is growing preference for multi-scale graphs \cite{li2020dynamic,dang2021msr,zhou2021learning}.
DMGNN \cite{li2020dynamic} uses dynamic multi-scale graphs to represent body segments at different scales. The segmentation is manually performed by averaging predefined joint groups. MSR-GCN \cite{dang2021msr} uses multilayer perceptrons to reduce or increase the number of joints to construct the multi-scale graph. The former lacks flexibility and the latter lacks clarity in terms of interpretability of the graph.

\paragraph{Graph Convolution}
In early approaches, a simple message passing is implemented depending on spatial adjacency to learn spatial-only dependencies \cite{mao2020history} with additional temporal CNNs (e.g., TCNs) \cite{lea2017temporal} employed to learn temporal-only dependencies \cite{cui2020learning,li2020dynamic}, which cannot effectively model spatio-temporal relations. Some recent work \cite{sofianos2021space,ma2022progressively} has begun to apply graph convolution to both spatial and temporal dependencies by additionally building temporal graphs over the trajectory of joints and implementing graph convolutions separately on the spatial and temporal graphs.
These approaches largely work in a static manner that shares the parameterized graph among all samples and cannot address diverse data-dependent relations. Different motion patterns underlying different individuals are often highly complex, highlighting the need for dynamic methods \cite{chen2021channel}. Another related work to ours is the dynamic routing \cite{sabour2017dynamic}, which defines dynamic aggregation based on routing-by-agreement.

What differentiates our work from existing approaches is that we build a dense graph with 4D adjacency modeling from a global perspective over the whole motion sequence instead of trajectories or poses. Another key distinction is that the proposed graph convolution is implemented dynamically depending on data-dependent relevance among nodes in the graph, whereas existing approaches implement graph convolution in a static manner depending on shared parameterized graphs among all samples.

\section{Problem Formulation and Preliminaries}

\subsection{Notations}
Suppose that a motion sequence consists of $T$ consecutive frames, in which each pose has $M$ joints. Then the motion sequence is $\tens{X}_{1:T} = [\mathbf{X}_1, \mathbf{X}_2, \cdots, \mathbf{X}_T] \in \mathbb{R}^{T\times M \times D}$, where $\mathbf{X}_t \in \mathbb{R}^{M\times D}$ represents the pose at time $t$, and each joint has $D$ features. Given the observed motion sequence $\tens{X}_{1:T_\text{h}}$ with $T_\text{h} < T$, the human motion prediction task is to accurately generate the future motion sequence $\tens{X}_{T_\text{h}+1:T}$.

Formally, a graph is a triplet $G = (\mathcal{V}, \mathcal{E}, W)$, where $\mathcal{V} = \{1,2,\cdots,N\}$ is a set of $N$ nodes, $\mathcal{E} \subseteq \mathcal{V} \times \mathcal{V}$ is a set of edges defined as order pairs $(n,m)$, and $W: \mathcal{E}\to\mathbb{R}$ is a map from the set of edges to scalars $w$, with $w_{nm}$ denoting the weight of $(n,m)$. The neighborhood $\mathcal{N}_G(m)$ of a node $m$ is a subset of $\mathcal{V}$ composed of $m$'s adjacent nodes.
The graph can be specified by its adjacency matrix $\mathbf{A} \in \mathbb{R}^{N\times N}$, with $A_{mn}$ specifying the edge from node $n$ to node $m$.

\subsection{Neural Message Passing}

During each message passing layer, the hidden representation of each node is updated according to information aggregated from its neighborhood. Let $\mathbf{F}\in\mathbb{R}^{N\times D}$ denote the node feature matrix, where $\mathbf{f}_i \in \mathbb{R}^{D}$ is the $D$-dimensional feature vector of node $i$. This message-passing update operation can be expressed as the following two steps:
\begin{equation}\label{message_passing}
\begin{split}
    & \mathbf{m}_i^{(k)} = \textsc{aggregate}(\{\mathbf{h}_j^{(k-1)} \vert j \in\mathcal{N}_G(i) \}), \\
    & \mathbf{h}_i^{(k)} = \textsc{update}(\mathbf{h}_i^{(k-1)}, \mathbf{m}_i^{(k)}),
\end{split}
\end{equation}
where $\mathbf{h}_i^{(k)}$ is the representation of node $i$ at layer $k$ and $\mathbf{h}_i^{(0)}=\mathbf{f}_i$. \textsc{aggregate} is a differentiable function that maps a set of vectors to a single vector, and the update function combines the message $\mathbf{m}_i^{(k)}$ aggregated from node $i$'s neighborhood with $\mathbf{h}_i^{(k-1)}$ to generate the updated representation.
Many message passing frameworks rely on linear operations followed by an element-wise non-linearity \cite{hamilton2020graph}, which can be performed efficiently by matrix operations as follows:
\begin{equation}
	\mathbf{H}^{(k)} = \textsc{propagate}(\mathbf{H}^{(k-1)}, \mathbf{A}^{(k)}; \theta^{(k)}),
\end{equation}
where $\mathbf{H}^{(0)}=\mathbf{F}$ and \textsc{propagate} is a propagation function depending on the adjacency matrix $\mathbf{A}$ and parameters $\theta^{(k)}$.

\begin{figure*}[!ht]
\centering
\includegraphics[width=0.99\textwidth]{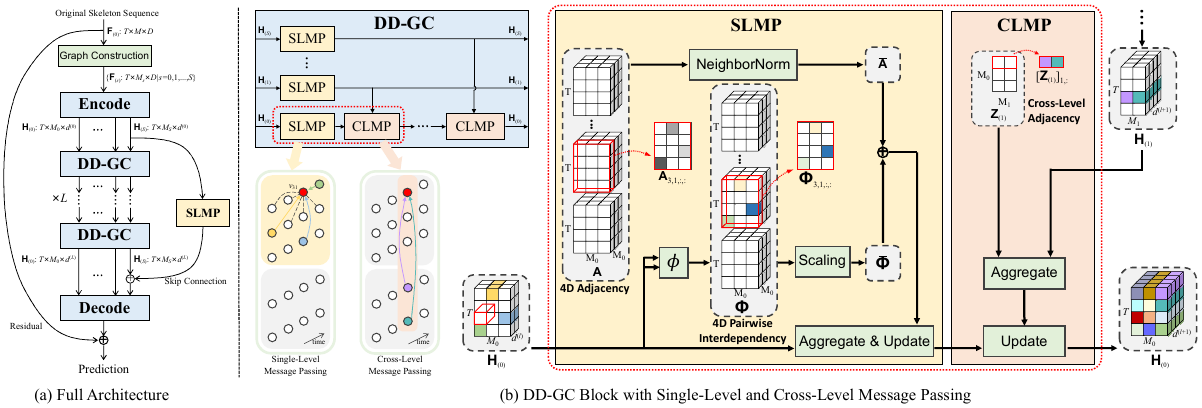} 
\caption{\textbf{Network architecture of our DD-GCN}. (a) Full architecture overview. The number of DD-GC blocks ($L$) and levels of message passing ($S$) are hyperparameters that can be adjusted for performance/complexity trade-off. (b) DD-GC block explained in detail. A dynamic dense graph convolution (DD-GC) block deploys a multi-pathway design to perform single-level (SLMP) and cross-level (CLMP) message passing over the graph with 4D adjacency.}
\label{network}
\end{figure*}

\section{Dense Graph Construction}
An intuitive approach adopted in many existing approaches is to build graphs depending on spatial adjacency in the pose, such that the motion sequence is characterized by a series of disjoint pose graphs along the time axis. In this way, features are aggregated using GCNs and then learned using temporal CNNs (e.g., TCNs). We argue that such separate feature learning is not sufficient to capture the complex spatio-temporal dynamic patterns underlying the motion sequence.

\subsection{Single-Level Dense Graph with 4D Adjacency}
Consider the general case where the motion sequence is modeled by a single dense graph $G_\text{st}=(\mathcal{V}_\text{st}, \mathcal{E}_\text{st}, W_\text{st})$. The node set $\mathcal{V}_\text{st}$ is composed of $T\cdot M$ nodes representing all the joints in the motion sequence, where node $v_{ij}$ stands for the $j$-th joint at time $i$. An edge $(v_{mn}, v_{ij})$ connects $v_{mn}$ and $v_{ij}$ if they are related by the spatio-temporal relationship that characterizes the motion. $W_\text{st}: \mathcal{E}_\text{st}\to\mathbb{R}$ is a map from the edge set to weight values $w$, with $w_{mnij}$ the weight of $(v_{mn}, v_{ij})$. The spatio-temporal neighborhood of a node $v_{ij}$ is defined as $\mathcal{N}_\text{st}(v_{ij}) = \{v_{mn}\vert(v_{mn},v_{ij})\in\mathcal{E}_\text{st}\}$. The graph can be fully specified by a four-dimensional tensor $\tens{A}\in\mathbb{R}^{T\times M \times T\times M}$, whose element-wise definition is:
\begin{equation}
    \tens{A}_{i,j,m,n} = \begin{cases}
    w_{mnij}, & \text{if $(v_{mn}, v_{ij})\in\mathcal{E_\text{st}}$};   \\
    0, & \text{otherwise}.
    \end{cases}
\end{equation}
$\tens{A}$ can be reshaped to a square matrix $\mathbf{A} \in \mathbb{R}^{TM\times TM}$ that can be interpreted as a block matrix:
\begin{equation}\label{block_matrix}
\mathbf{A} = \left[ \begin{array}{cccc}
     \mathbf{A}_{11} & \mathbf{A}_{12} & \cdots & \mathbf{A}_{1T} \\
     \mathbf{A}_{21} & \mathbf{A}_{22} & \cdots & \mathbf{A}_{2T} \\
     \vdots & \vdots & \ddots & \vdots \\
     \mathbf{A}_{T1} & \mathbf{A}_{T2} & \cdots & \mathbf{A}_{TT}
\end{array} \right],
\end{equation}
where $\mathbf{A}_{im}=\tens{A}_{i,:,m,:}\in\mathbb{R}^{M\times M}$ is a square block with element $(A_{im})_{jn}=\tens{A}_{i,j,m,n}$. The main-diagonal blocks specify the spatial adjacency in different poses, and the main-diagonal of all blocks specify the temporal adjacency in different trajectories.
The node features are denoted by a three-dimensional tensor $\tens{F}\in\mathbb{R}^{T\times M\times D}$, where $\mathbf{F}_{i}=\tens{F}_{i,:,:}\in\mathbb{R}^{M\times D}$ denotes the pose feature matrix at time $i$, with each node corresponding to a row vector in the matrix such that $\mathbf{f}_{ij}=\tens{F}_{i,j,:}\in\mathbb{R}^D$ is the feature vector of node $v_{ij}$.

\subsection{Adaptive Dense Graph Construction}

To exploit the prior knowledge of human skeletons, we can specify the dense graph depending on natural connectivity. That is, an edge connects two nodes at the same frame or across two different frames if their corresponding joints are connected by a bone in the skeleton. However, such graph is a low-level representation of human motion, in the sense that it involves the low-level segmentation of the human body into bones connected by joints. However, there are various types of motions exhibited by the human body, so much so that a dense graph alone does not suffice. To address this challenge, we propose to construct the dense graph based on human anatomy to model human motion at both high and low levels.

Given the dense graph $G_{(0)}=(\mathcal{V}_{(0)}, \mathcal{E}_{(0)}, W_{(0)})$ that describes motion patterns at the finest scale, we derive from it a series of $S$ dense graphs at different coarser scales, denoted $G_{(s)}=(\mathcal{V}_{(s)}, \mathcal{E}_{(s)}, W_{(s)})$ for all $s=1,2,\cdots,S$. Note that we omit the subscript ``st'' for notational simplicity. The dense graph $G_{(s)}$ is constructed in an end-to-end fashion by learning a \textbf{skeleton transformation matrix} $\mathbf{Z}_{(s)}\in\mathbb{R}^{M\times M_s}$ over the nodes in $G_{(0)}$ that groups them into $M_s$ parts (Fig. \ref{skeleton_transform}).
Formally, let $\left[\mathbf{F}_{(0)}\right]_i\in\mathbb{R}^{M\times D}$ denote the pose feature matrix in $G_{(0)}$ at any given time $i$. Then we compute the pose features in the coarser graph $G_{(s)}$ as follows:
\begin{equation}\label{skl_trans}
    \left[\mathbf{F}_{(s)}\right]_i = \mathbf{Z}_{(s)}^\top \left[\mathbf{F}_{(0)}\right]_i\in\mathbb{R}^{M_s\times D}, s=1,2,\cdots,S.
\end{equation}
We set $\mathbf{Z}_{(s)}$ as trainable parameters with predefined initial values based on the anatomical structure. For cross-level learning, we build a series of \textbf{cross-level graphs} that are bipartite graphs.
The overall dense graph is formed from all single-level dense graphs and cross-level graphs, describing the diverse and complex spatio-temporal motion patterns at different levels of abstraction. Its unique representational capacity enables more precise and effective feature learning.

\subsection{Induced Local Subgraph}
Given a dense graph $G_\text{st}=(\mathcal{V}_\text{st}, \mathcal{E}_\text{st}, W_\text{st})$ and its corresponding $\mathbf{A}$, we define the \textbf{pose subgraph} corresponding to each pose as an induced subgraph of $G_\text{st}$ formed from all nodes representing the pose and all the edges connecting them. The pose subgraph models the human skeleton pose in the spatial domain.
Formally, the pose subgraph at time $i$ is denoted by $\hat{G}_i = (\hat{\mathcal{V}}_i, \hat{\mathcal{E}}_i, \hat{W}_i)$, with $\hat{\mathcal{V}}_i = \{v_{ij}\vert j=1,2,\cdots,M\}$. Then the dense graph $G_\text{st}$ is partially characterized by a sequence of $T$ disjoint pose subgraphs $\hat{G}_1,\hat{G}_2,\cdots,\hat{G}_T$, with their node sets and edge sets satisfying:
$\mathcal{V}_\text{st} = \hat{\mathcal{V}}_1 \cup \hat{\mathcal{V}}_2 \cup \cdots \cup \hat{\mathcal{V}}_T; \mathcal{E}_\text{st} \supsetneqq \hat{\mathcal{E}}_1 \cup \hat{\mathcal{E}}_2 \cup \cdots \cup \hat{\mathcal{E}}_T.$
Note that we use a block matrix $\mathbf{A}$ as in Eq. (\ref{block_matrix}) corresponding to $G_\text{st}$. Then each pose subgraph $\hat{G}_i$ is fully specified by the main-diagonal block $\mathbf{A}_{ii} = \tens{A}_{i,:,i,:}$ for all $i=1,2,\cdots,T$. And the off-diagonal blocks of $\mathbf{A}$ specify the cross-pose connections between two distinctive poses.

Extrapolated from pose subgraphs in the spatial domain, the \textbf{trajectory subgraph} is an induced subgraph of $G_\text{st}$ defined as $\tilde{G}_j = (\tilde{\mathcal{V}}_j, \tilde{\mathcal{E}}_j, \tilde{W}_j)$ corresponding to the $j$-th joint, with $\tilde{\mathcal{V}}_j = \{v_{ij}\vert i=1,2,\cdots,T\}$. There are a total of $M$ disjoint trajectory subgraphs that model the motion sequence in the time domain. Analogous to pose subgraphs, trajectory subgraphs are specified by the main-diagonal of all the blocks in $\mathbf{A}$, i.e., $\tens{A}_{:,j,:,j}$ for all $j=1,2,\cdots,M$.
These initial conceptual specifications presented in this section are meant to facilitate the next steps of implementing the graph convolution.

\section{Dynamic Dense Graph Convolutional Network}

In above sections, we have discussed the general neural message passing framework in a relatively abstract fashion, which involves \textsc{aggregate} and \textsc{update} functions.
In this subsection we give concrete instantiations to these functions in order to implement the proposed DD-GCN. Note that we will omit the subscript ``st'' and superscript denoting the layer for notational simplicity when necessary.
In this work, we propose a dynamic message passing framework, wherein we employ the aggregator to dynamically learn from input data and generate messages based on sample-specific relevance among nodes to reflect the underlying distinctive relations. And we apply joint-specific projections for node-level update accordingly. In contrast, previous approaches simply take weighted average of the neighbor vectors, where the weights are predefined or parameterized as edge attributes shared by all samples, and the update is implemented using a shared projection matrix among all nodes.

\subsection{Dynamic Message Passing}
The aggregation step targets on generating informative messages from neighborhoods to facilitate the following update. The aggregation is in essence a set function that maps a set of representation vectors in the neighborhood to a vector.
Formally, the \textsc{aggregate} function is defined as:
\begin{equation}\label{aggregation}
\begin{split}
    \mathbf{m}_{ij} & = \textsc{aggregate}(\{\mathbf{h}_{mn} \vert v_{mn} \in\mathcal{N}(v_{ij})\} \cup \{\mathbf{h}_{ij}\}) \\
    & = \sum_{v_{mn}\in\mathcal{N}(v_{ij})} \big( \alpha_{ijmn} + \frac{1}{\sqrt{d_\text{out}}}\phi(\mathbf{h}_{mn}, \mathbf{h}_{ij}) \big)\mathbf{h}_{mn},
\end{split}
\end{equation}
where $d_\text{out}$ is the dimension of the final output of this layer, $\alpha_{ijmn}$ is a shared weight for all samples, and $\phi(\mathbf{h}_{mn}, \mathbf{h}_{ij})$ is a data-dependent weight inferred in the forward pass. The initial node representation at layer 0 is the feature vector, i.e., $\mathbf{h}_{ij}^{(0)}=\mathbf{f}_{ij}=\tens{F}_{i,j,:}$. The shared term $\alpha_{ijmn}$ is set as trainable parameters to reflect generic common patterns:
\begin{equation}
    \alpha_{ijmn} = \frac{1}{\vert\mathcal{N}(v_{ij})\vert}\tens{A}_{i,j,m,n} = \frac{1}{\vert\mathcal{N}(v_{ij})\vert}w_{mnij},
\end{equation}
where $\vert\cdot\vert$ denotes the cardinality of a set.
The data-dependent term $\phi(\mathbf{h}_{mn}, \mathbf{h}_{ij})$ is designed to model the characteristic relation among nodes for each sample, with $\phi:\mathbb{R}^{d_\text{in}}\times\mathbb{R}^{d_\text{in}}\to\mathbb{R}$ a map from two vectors to a scalar. Specifically, we design two nonlinear weight functions $\phi_1$ and $\phi_2$ as follows:
\begin{equation}
\begin{split}
    \phi_1 &= \text{softsign}\big(\text{mean}(\mathbf{h}_{ij}) - \text{mean}(\mathbf{h}_{mn})\big); \\
    \phi_2 &= \text{softsign}(-\mathbf{h}_{ij}^\top \mathbf{h}_{mn}). \\
\end{split}
\end{equation}
We apply a scaling operation to the data-dependent weight which is to divide it by $\sqrt{d_\text{out}}$. By doing so it is scaled down to values close to $\alpha_{ijmn}$ in case it grows dominantly large such that the shared weight end up useless for message generating. Scaling down the weights can mitigate the vanishing gradients problem in the upcoming update step by keeping them away from regions where the gradient might vanish.


Put simply, in the aggregation step we refine the mean aggregator commonly adopted in existing approaches such that it takes the data-dependent weighted average of the neighbors to generate distinctive messages for update. The \textsc{update} function in DD-GCN takes as input the message from the neighborhood and the node's previous representation and generates the new updated representation, which is defined as:
\begin{equation}\label{update}
\begin{split}
    \mathbf{h}_{ij} & = \textsc{update}(\mathbf{h}_{ij}, \mathbf{m}_{ij}) \\
    & = \sigma\big(\mathbf{\Theta}_j (\mathbf{h}_{ij} + \mathbf{m}_{ij}) + \mathbf{b}\big),
\end{split}
\end{equation}
where $\sigma$ denotes an element-wise non-linearity (e.g., tanh), $\mathbf{b}\in\mathbb{R}^{d_\text{out}}$ is a bias term, and $\mathbf{\Theta}_j\in\mathbb{R}^{d_\text{out}\times d_\text{in}}$ is the trainable joint-specific projection matrix corresponding to the $j$-th joint.

\subsection{Adaptive Multi-Level Message Passing}

On top of the basic message passing framework introduced in the previous paragraph, we design a multi-pathway framework with light-weight lateral connections to capture and fuse informative features at and across different levels. Given $G_{(0)}$ and the corresponding feature vector $\left[\mathbf{f}_{(0)}\right]_{ij}$ of node $v_{ij}$, we can obtain the node feature $\left[\mathbf{f}_{(s)}\right]_{ij}$ in $G_{(s)}$ for any $s=1,2,\cdots,S$ using the skeleton transformation $\mathbf{Z}_{(s)}$ as in Eq. (\ref{skl_trans}).
In each pathway, we first encode the input node features $\left[\mathbf{f}_{(s)}\right]_{ij}$ into hidden representations $\left[\mathbf{h}_{(s)}\right]_{ij}$ and then learn them using the DD-GC layers discussed above.
Moreover, we interleave the pathway at scale $0$ with multiple light-weight cross-level message passing modules after every stage to imbue it with information from other levels as guidance for feature learning at the original level.
Given the pose representation $\left[\mathbf{H}_{(s)}\right]_i\in \mathbb{R}^{M_s\times d_s}$ at any given time $i$, the cross-level message passing module takes it as input and generates a cross-level message $\left[\mathbf{H}_{(s,0)}\right]_i\in\mathbb{R}^{M\times d}$ through graph convolution depending on the cross-level graph:
\begin{equation}
    \left[\mathbf{H}_{(s,0)}\right]_i = \mathbf{\bar{Z}}_{(s)}\mathbf{A}_{(s)}\left[\mathbf{H}_{(s)}\right]_i \mathbf{\Theta}_{(s)},
\end{equation}
where $\mathbf{\Theta}_{(s)}\in\mathbb{R}^{d_s\times d}$ is the trainable projection matrix, $\mathbf{A}_{(s)}\in\mathbb{R}^{M_s\times M_s}$ is the trainable adjacency matrix initialized as $\mathbf{A}_{(s)}=\mathbf{I}$, and $\mathbf{\bar{Z}}_{(s)}\in\mathbb{R}^{M\times M_s}$ is the \textbf{inverse skeleton transformation matrix} defined as:
\begin{equation}
    \mathbf{\bar{Z}}_{(s)} = \mathbbm{1}(\mathbf{Z}_{(s)} \neq 0).
\end{equation}
The projection matrix $\mathbf{\Theta}_{(s)}$ is used only if the feature dimension at the two levels doesn't match. The cross-level message passing modules are made light-weight by sharing projection and adjacency matrices across different poses. The cross-level representations are then fused with the representation at the original level from the previous stage:
\begin{equation}
\begin{split}
    \left[\mathbf{h}_{(0)}\right]_{ij} & = \textsc{update}_\text{cross-level}(\left[\mathbf{h}_{(0)}\right]_{ij}, \sum_{s=1}^S \left[\mathbf{h}_{(s,0)}\right]_{ij}) \\
    & = \left[\mathbf{h}_{(0)}\right]_{ij} + \sum_{s=1}^S \left[\mathbf{h}_{(s,0)}\right]_{ij}.
\end{split}
\end{equation}
For the sake of notational simplicity, we use $\left[\mathbf{h}_{(0,0)}\right]_{ij}$ to denote $\left[\mathbf{h}_{(0)}\right]_{ij}$ and rewrite the equation above as:
\begin{equation}
    \left[\mathbf{h}_{(0)}\right]_{ij} = \sum_{s=0}^S \left[\mathbf{h}_{(s,0)}\right]_{ij}.
\end{equation}

\begin{figure}[t]
\centering
\includegraphics[width=0.99\columnwidth]{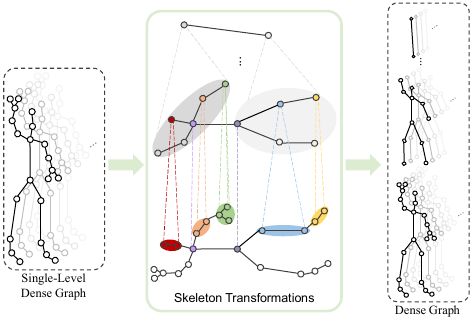} 
\caption{\textbf{Dense graph construction}. The input skeleton sequence corresponds to the single-level dense graph with nodes representing joints throughout the whole sequence and edges connecting node pairs, that is specified by a 4D adjacency matrix. The dense graph is constructed via learnable skeleton transformations that group joints into semantics-adjacent parts.}
\label{skeleton_transform}
\end{figure}

\subsection{Dynamic Dense Graph Convolution}
In the preceding paragraphs we present in-depth node-level definitions of the message passing framework at and across different levels.
We unify them and present the overall dynamic dense graph convolution (DD-GC) in tensor notation. For notational clarity we will use superscripts to denote levels and omit the layer indices. The proposed DD-GC is defined as:
\begin{equation}
\begin{split}
    \tens{H}^{(s)}_{i,j,:} &= \sigma \Big( \boldsymbol{\mathsf{\Theta}}^{(s)}_{j,:,:} \big( \sum_{m=1}^T\sum_{n=1}^M (  \tens{\bar{A}}^{(s)}_{i,j,m,n}+\boldsymbol{\mathsf{\bar{\Phi}}}^{(s)}_{i,j,m,n} ) \tens{H}^{(s)}_{m,n,:} \\ & + \tens{H}^{(s)}_{i,j,:}\big) + \tens{b} \Big), s=0,\cdots,S;\\
    \tens{H}^{(0)}_{i,:,:} &= \sum_{s=0}^S \mathbf{\bar{Z}}^{(s)}\mathbf{A}^{(s)}\tens{H}^{(s)}_{i,:,:},
\end{split}
\end{equation}
where $\tens{\bar{A}}$ and $\boldsymbol{\mathsf{\bar{\Phi}}}$ are the normalized $\tens{A}$ and the scaled $\boldsymbol{\mathsf{\Phi}}$ respectively, with $\tens{\bar{A}}_{i,j,m,n}=\frac{1}{\vert\mathcal{N}(v_{ij})\vert}\tens{A}_{i,j,m,n};\boldsymbol{\mathsf{\bar{\Phi}}}_{i,j,m,n}=\frac{1}{\sqrt{d_\text{out}}}\phi(\tens{H}^{(s)}_{m,n,:},\tens{H}^{(s)}_{i,j,:}).$
In the case of $s=0$, we set $\mathbf{\bar{Z}}^{(s)}$ and $\mathbf{A}^{(s)}$ as identity matrices. The initial representation at layer 0 is the input node features, i.e.,  $\tens{H}^{(0)}=\tens{F}$.

\section{Network Architecture}

The overall architecture of the proposed DD-GCN is illustrated in Fig. \ref{network}. The network first takes the original skeleton sequence as input, constructs the dense graph, and extract the corresponding features at different levels. Then the graph convolution is performed using a stack of $L$ dynamic dense graph convolution (DD-GC) blocks.
In each block, the features are first aggregated and updated using single-level message passing (SLMP) blocks. And then the cross-level message passing (CLMP) blocks propagate the features across different pathways. The features at each single level are learned in a dynamic way that depends on the shared adjacency as well as sample-specific relevance among nodes in the graph. The features from different levels are propagated over the bipartite graph with cross-level adjacency.
In implementation, we utilize three ($L=3$) DD-GC blocks and three ($S=2$) levels of graph representation in total. These two hyperparameters are situational for model performance/complexity trade-off. Through skeleton transformation (Fig. \ref{skeleton_transform}), the numbers of joints or body parts in the three levels are set $22,11,2$, respectively,  which are obtained by learning the corresponding skeleton transformations in the graph construction step at the very beginning of the network. Moreover, we apply a global residual connection, as commonly adopted in previous work. And we apply an additional skip connection in the pathway at the highest level, which tries to directly preserve information from previous rounds of message passing using a SLMP block.

\subsection{Single-Level Message Passing}
In each single-level message passing (SLMP) block, features are learned in a dynamic way that depends on the shared adjacency as well as sample-specific relevance among nodes in the graph. A SLMP block consists of graph convolution, batch normalization, dropout and activation. Table \ref{slmp} presents structures of first three SLMP blocks in DD-GCN.
\begin{table}[t]
\centering
\caption{The structure of SLMP blocks at different levels in DD-GCN. The size of adjacency and features are denoted by $[T,M_s,T,M_s]$ and $[N,T,M_s,D]$ respectively.}
\label{slmp}
\resizebox{\columnwidth}{!}{
\begin{tabular}{c|c|c|c} \hline
		level & operations & adjacency  & feature  \\ \hline
		\multirow{3}{*}{1}  & aggregate & [50,22,50,22] & [16,50,22,3]\\ \cline{2-4}
		\multirow{3}{*}{}   & update & - & [16,50,22,128]\\ \cline{2-4}
		\multirow{3}{*}{ }  & bn, dropout, tanh & - & [16,50,22,128]  \\ \hline
		
		\multirow{3}{*}{2}  & aggregate & [50,11,50,11] & [16,50,11,3]\\ \cline{2-4}
		\multirow{3}{*}{}   & update & - & [16,50,22,128]\\ \cline{2-4}
		\multirow{3}{*}{ }  & bn, dropout, tanh & - & [16,50,11,128]  \\ \hline
		
		\multirow{3}{*}{3}  & aggregate & [50,2,50,2] & [16,50,2,3]\\ \cline{2-4}
		\multirow{3}{*}{}   & update & - & [16,50,22,128]\\ \cline{2-4}
		\multirow{3}{*}{ }  & bn, dropout, tanh & - & [16,50,2,128]  \\ \hline
\end{tabular}}
\end{table}

\subsection{Cross-Level Message Passing}
In each cross-level message passing (CLMP) block, the features from different levels are propagated over the bipartite graph with cross-level adjacency. We present an exemplary case of CLMP from level 1 to level 0 in Table \ref{clmp}.
\begin{table}[t]
	\centering
	\caption{The structure of an exemplary CLMP block between level 1 and level 0 in DD-GCN.}
	\label{clmp}
	\resizebox{\columnwidth}{!}{
	\begin{tabular}{c|c|c|c|c} \hline
			step & operations & adjacency & feature & output  \\ \hline
			1  & aggregate & [22,11] & $\tens{H}_{(1)}$ [16,50,11,128] & [16,50,22,128] \\\hline
			2  & update & - & $\tens{H}_{(0)}$ [16,50,22,128] & [16,50,22,128] \\ \hline
	\end{tabular}}
	
\end{table}

\subsection{An Alternative for Message Passing}
Our idea of the dynamic message passing framework is generic, in the sense that it can be instantiated with different implementation specifications. Considering both the accuracy and the computational cost, we provide an alternative desgin of the single-level message passing block, in which the shared 4D adjacency is replaced with two 3D adjacency tensors, each containing a proportion of the 4D adjacency tensor’s elements corresponding to edges in the pose subgraph and trajectory subgraph respectively. It can be regarded as a special case of the original SLMP block in DD-GCN where we mask certain elements in the 4D adjacency tensor. Table \ref{slmp_alternative_edited} shows the structure of such edition of the SLMP block.
\begin{table}[t]
	\centering
	\caption{The structure of alternative SLMP with 3D adjacency over the pose subgraph and trajectroy subgraph.}
	\label{slmp_alternative_edited}
	\resizebox{\columnwidth}{!}{
		\begin{tabular}{c|c|c|c} \hline
			step & operations & adjacency & feature \\ \hline
			1 & pose aggregate & [50,22,22] & [16,50,22,3] \\ \hline
			2 & trajectory aggregate & [22,50,50] & [16,50,22,3] \\ \hline
			3 & update & - & [16,50,22,128] \\ \hline
			4 & bn, dropout, tanh & - & [16,50,22,128] \\ \hline
	\end{tabular}}
	
\end{table}

    \begin{table*}[t]
        \centering
        \caption{Average MPJPEs (Mean Per Joint Position Errors) in millimeter for normal and extended long-term prediction on H3.6M dataset. The baseline methods include ERD~\cite{fragkiadaki2015recurrent}, Res-RNN~\cite{martinez2017human}, Skel-TNet~\cite{guo2019human}, convSeq2Seq~\cite{li2018convolutional}, LTD~\cite{mao2019learning}, MSR-GCN~\cite{dang2021msr}, DMGNN~\cite{li2020dynamic}, HisRep~\cite{mao2020history}, PGBIG~\cite{ma2022progressively}, siMLPe~\cite{guo2023back} and BSTG-Trans~\cite{mo2023bstg}. Note that the evaluation provided in these papers differs in validation and testing strategies. Therefore we re-trained the models separately for different settings and evaluate them following the same paradigm as previous state-of-the-art methods \cite{mao2019learning,li2020dynamic,martinez2017human}.}
        \label{avg_long_pred}
        \resizebox{\textwidth}{!}{
        \begin{tabular}{ll|l|cc|cccccc} \hline
        \multirow{2}{*}{Model} & \multirow{2}{*}{Venue} & \multirow{2}{*}{Backbone} & \multicolumn{2}{c|}{Normal} & \multicolumn{6}{c}{Extended} \\ 
         &  &  & 560ms & 1000ms & 1200ms&1300ms & 1400ms & 1600ms & 1800ms & 2000ms \\ \hline
         ERD~\cite{fragkiadaki2015recurrent} & ICCV'15 &  &91.32& 128.37    & 140.91&144.65& 151.89& 160.22& 164.99& 181.05 \\
         Res-RNN~\cite{martinez2017human}  & CVPR'17 & & 86.12& 122.45    & 135.24&148.62& 146.61& 153.92& 159.88& 174.61 \\
         Skel-TNet~\cite{guo2019human}& AAAI'19 & \multirow{-3}{*}{RNN} &93.80&  124.3   &  138.74&144.30& 152.48& 153.64& 168.42& 179.69 \\ \hdashline
         convSeq2Seq~\cite{li2018convolutional}& CVPR'18 & CNN &82.18 &117.44     &134.85 &141.57&145.23 &148.83 &159.29 &171.66 \\  \hdashline
         LTD~\cite{mao2019learning} & ICCV'19 &  & 80.75 & 112.71      &125.38&127.84 &132.21 &137.90 &144.66 &154.40 \\
         MSR-GCN~\cite{dang2021msr} & ICCV'21 &  & 77.72 & 112.18      &129.50&135.45 &138.33 &144.90 &151.98 &159.05\\
         DMGNN~\cite{li2020dynamic} & CVPR'20 &  &81.18 &116.44  &133.85 &139.37&144.23 &147.83 &155.29 &166.66 \\
         HisRep~\cite{mao2020history} & ECCV'20 &  &79.18& 114.44     &131.85 &136.61&142.23 &145.83& 153.29& 164.66 \\
         PGBIG~\cite{ma2022progressively} & CVPR'22 & \multirow{-5}{*}{GCN} & 77.83 & 111.95    &122.44 &127.13&129.53 &134.98 &142.42 &152.59 \\ \hdashline
         siMLPe~\cite{guo2023back}  & WACV'23 & MLP-Mixer &79.82& 114.07    & 126.24&128.90& 135.18& 138.97& 145.71& 154.44 \\ \hdashline
         BSTG-Trans~\cite{mo2023bstg} & TMM'23 & Transformer &79.63&112.34  &125.64  &127.96 &132.86 &137.56 & 143.20 &  153.97 \\ \hline
         Ours (Single Level)	& & & 79.10&\uline{110.28}&\uline{122.56} &\uline{126.08}&\uline{129.54}&\uline{134.10}&\uline{141.21}&\uline{151.99}\\ 
         Ours & & \multirow{-2}{*}{GCN} & \uline{78.82} & \textbf{110.07} &\textbf{122.24}   &\textbf{125.78}&130.18 &\textbf{133.97} &\textbf{140.71} &\textbf{150.44} \\ \hline
        \end{tabular}}
        \vspace{-2em}
    \end{table*}

    \begin{table}[!t]
        \centering
        \caption{Average MPJPEs (Mean Per Joint Position Errors) in millimeter on AMASS dataset.}
        \label{tab:amass}
        \resizebox{\columnwidth}{!}{
        \begin{tabular}{c|cccccccc} \hline
        \multirow{2}{*}{Model} & \multicolumn{8}{c}{Mean Per Joint Position Error (mm)} \\
           & 80ms & 160ms & 320ms & 400ms & 560ms & 720ms & 880ms & 1000ms \\ \hline
        LTD 	&11.0 	&20.7 	&37.8 	&45.3 	&57.2 	&65.7 	&71.3 	&75.2\\
        HisRep	&11.3 	&20.7 	&35.7 	&42.0 	&51.7 	&58.6 	&63.4 	&67.2\\
        siMLPe  &10.8 	&19.6 	&34.3 	&40.5 	&50.5 	&57.3 	&62.4 	&65.7\\ 
        Ours	&\textbf{10.4} &\textbf{19.1} &\textbf{33.6} &\textbf{39.8} &\textbf{49.3} &\textbf{56.5} &\textbf{61.3} &\textbf{64.6}
        \\ \hline
        \end{tabular}}
        \vspace{-1em}
    \end{table}
    
    \begin{table}[!t]
        \centering
        \caption{Average MPJPEs (Mean Per Joint Position Errors) in millimeter on 3DPW dataset.}
        \label{tab:3dpw}
        \resizebox{\columnwidth}{!}{
        \begin{tabular}{c|cccccccc} \hline
        \multirow{2}{*}{Model} & \multicolumn{8}{c}{Mean Per Joint Position Error (mm)} \\
            & 80ms & 160ms & 320ms & 400ms & 560ms & 720ms & 880ms & 1000ms \\ \hline
        LTD &12.8&23.5&39.8&46.4&57.7&65.5&71.6&76.2\\
        HisRep & 12.9&23.2&39.5&46.0&57.1&64.8&71.2&75.9\\
        siMLPe  &12.6&22.8&38.9&45.3&56.0&63.6&69.9&74.6\\ 
        Ours	&\textbf{11.7}&\textbf{22.0}&\textbf{38.1}&\textbf{44.6}&\textbf{55.1}&\textbf{62.5}&\textbf{68.2}&\textbf{72.3}\\ \hline
        \end{tabular}}
        \vspace{-1em}
    \end{table}

    \begin{table}[t]
        \centering
        \caption{Average MPJPEs (Mean Per Joint Position Errors) in millimeter on NTU RGB+D dataset.}
        \label{tab:ntu}
        \resizebox{\columnwidth}{!}{
        \begin{tabular}{c|cccccccc} \hline
        frame & 1 & 3 & 5 & 7 & 9 & 11 & 13 & 15 \\ \hline
        LTD    & 26.0 & 45.2 & 58.4 & 69.1 & 78.9 & 87.8 & 92.3 & 97.9 \\
        HisRep & 25.9 & 44.6 & 57.3 & 68.4 & 78.5 & 87.0 & 91.8 & 97.2 \\
        siMLPe & 26.4 & 44.9 & 57.3 & 68.0 & 77.9 & 85.3 & 90.1 & 96.7 \\
        Ours   & \textbf{23.2} & \textbf{41.7} & \textbf{54.4} & \textbf{64.5} & \textbf{72.8} & \textbf{79.7} & \textbf{85.4} & \textbf{90.2} \\ \hline
        \end{tabular}}
        \vspace{-1em}
    \end{table}

\subsection{Training}
To train our model, we make use of the Mean Per Joint Position Error (MPJPE) \cite{ionescu2013human3}. Given the history motion sequence, the model is expected to generate future motion sequence that is as close as possible to the corresponding ground truth. Let $\tens{X}_n$ be the $n$-th output sample generated by the model and $\tens{\hat{X}}_n$ be the corresponding ground truth. Then the loss function for a total of $N$ training samples is defined as:
\begin{equation}
	\mathcal{L} = \frac{1}{NTM}\sum_{n=1}^N \sum_{i=1}^T \sum_{j=1}^M \lVert (\tens{\hat{X}}_{i,j,:})_n - (\tens{X}_{i,j,:})_n \rVert_2.
\end{equation}

\section{Analysis of Graph Convolution}
Since the trailblazing work of LTD \cite{mao2019learning}, different flavors of graph convolutions have been proposed in human motion prediction. We analyze them by reformulating them into a unified form and comparing them with ours in terms of feature aggregation and update. We further show that the proposed DD-GC is a generalization of them.

\subsection{Graph Convolution on Subgraph}
Recall that we use a dense graph to represent the whole skeleton sequence, with a 4D tensor $\tens{A}\in\mathbb{R}^{T\times M\times T\times M}$ specifying the adjacency of it, and that the 4D tensor can be reshaped to a block matrix $\mathbf{A}\in\mathbb{R}^{TM\times TM}$ (Eq. (\ref{block_matrix})).

\paragraph{Spatial Graph Convolution on Pose Subgraph}

As discussed in Section \textsc{IV}, pose subgraph $\hat{G}_i$ corresponding to the pose at $i$-th frame is fully specified by the main-diagonal block $\mathbf{A}_{ii}$ for $i=1,2,\cdots,T$.

The spatial graph convolution on pose subgraph is defined as the propagation of node representations within each pose, which can be performed by matrix operations as:
\begin{equation}
\begin{split}
	\mathbf{H}^{(k)} &= \textsc{propagate}_\text{s}(\mathbf{H}^{(k-1)}, \mathbf{A}^{(k)}; \theta^{(k)}) \\
	&= \sigma \big(   (\mathbf{A}^{(k)} \odot \mathbf{M}_\text{s}) \mathbf{H}^{(k-1)} \mathbf{\Theta}^{(k)} \big) \\
	&= \sigma(
	\left[ \begin{array}{cccc}
     \mathbf{A}_{11} & \mathbf{0} & \cdots & \mathbf{0} \\
     \mathbf{0} & \mathbf{A}_{22} & \cdots & \mathbf{0} \\
     \vdots & \vdots & \ddots & \vdots \\
     \mathbf{0} & \mathbf{0} & \cdots & \mathbf{A}_{TT}
    \end{array} \right]^{(k)}
    \mathbf{H}^{(k-1)} \mathbf{\Theta}^{(k)}),
\end{split}
\end{equation}
where $\mathbf{H}^{(k)}\in\mathbb{R}^{TM\times d^{(k)}}$ is the hidden representation at layer $k$, $\mathbf{\Theta}^{(k)}\in\mathbb{R}^{d^{(k-1)}\times d^{(k)}}$ is the trainable projection matrix, $\mathbf{M}_\text{s}$ is a fixed binary mask, and $\odot$ denotes the element-wise product. It can be easily seen that this is a special case of our DD-GC if we remove the term $\phi(\mathbf{h}_{mn},\mathbf{h}_{ij})$ in Eq. (\ref{aggregation}) and apply a mask to $\mathbf{A}$ such that all the off-diagonal blocks in $\mathbf{A}$ are zero matrices. In this case, the DD-GC becomes static spatial graph convolution, and the 4D adjacency reduces to a 3D adjacency of size $T\times M\times M$.
In an even more special case when the main-diagonal blocks are equal, i.e., $\mathbf{A}_{11}=\mathbf{A}_{22}=\cdots=\mathbf{A}_{TT}$, or equivalently, when all poses share the same spatial adjacency, the 3D adjacency reduces to a 2D adjacency of size $M\times M$, which is adopted in most of the previous work \cite{mao2019learning,dang2021msr,cui2020learning,li2020dynamic}.

\paragraph{Temporal Graph Convolution on Trajectory Subgraph}

Analogous to pose subgraphs, trajectory subgraph $\tilde{G}_j$ corresponding to the $j$-th joint is specified by a submatrix of $\mathbf{A}$ that is formed by taking the $j$-th element from every main diagonal of all the blocks in $\mathbf{A}$, which is equivalent to taking $\tens{A}_{:,j,:,j}$ and reshape it to a matrix.
The temporal graph convolution on trajectory subgraph can be formulated as:
\begin{equation}
\begin{split}
	\mathbf{H}^{(k)} &= \textsc{propagate}_\text{t}(\mathbf{H}^{(k-1)}, \mathbf{A}^{(k)}; \theta^{(k)}) \\
	&= \sigma \big(   (\mathbf{A}^{(k)} \odot \mathbf{M}_\text{t}) \mathbf{H}^{(k-1)} \mathbf{\Theta}^{(k)} \big) \\
	&= \sigma(
	\left[ \begin{array}{cccc}
     \mathbf{\Lambda}_{11} & \mathbf{\Lambda}_{12} & \cdots & \mathbf{\Lambda}_{1T} \\
     \mathbf{\Lambda}_{21} & \mathbf{\Lambda}_{22} & \cdots & \mathbf{\Lambda}_{2T} \\
     \vdots & \vdots & \ddots & \vdots \\
     \mathbf{\Lambda}_{T1} & \mathbf{\Lambda}_{T2} & \cdots & \mathbf{\Lambda}_{TT}
    \end{array} \right]^{(k)}
    \mathbf{H}^{(k-1)} \mathbf{\Theta}^{(k)}).
\end{split}
\end{equation}
This also is a special case of the DD-GC when we remove the dynamic term in Eq. (\ref{aggregation}) and apply the proper mask to the 4D adjacency such that every block in $\mathbf{A}$ is a diagonal matrix $\mathbf{\Lambda}_{im} = \text{diag}  (   \tens{A}_{i,1,m,1},\tens{A}_{i,2,m,2},\cdots,\tens{A}_{i,M,m,M}  )$. In this case, the DD-GC becomes temporal graph convolution, and the 4D adjacency reduces to a 3D adjacency of size $M\times T\times T$.
In a more special case when the elements of the diagonal in each block are equal, i.e., $\mathbf{\Lambda}_{im} = a_{im} \mathbf{I}$ where $a_{im}$ is a scalar, or equivalently, when all the trajectories share the same temporal adjacency, the 3D adjacency reduces to a 2D adjacency of size $T\times T$ that has elements $a_{im}$ for $i,m=1,2,\cdots,T$.

Earlier work usually employs temporal convolutions (temporal CNNs) for temporal modeling \cite{mao2019learning,cui2020learning,li2020dynamic}. In recent work they are replaced by the temporal graph convolution, which is usually employed in conjunction with the spatial one, in which case the graph convolution is performed separately on the spatial graph and the temporal graph \cite{ma2022progressively,sofianos2021space}. Nevertheless, these graph convolutions can all be treated as special cases of the DD-GC.

\subsection{Generalized Aggregation and Update}
We have presented the general form of message passing in Eq. (\ref{message_passing}) which involves an aggregation and an update step.

\paragraph{Neighborhood Aggregation}

The aggregation operations in most of the GCN-based methods in human motion prediction simply take the weighted sum or average of the neighbor representations to generate messages where the weights are set as trainable parameters. It is formulated as:
\begin{equation}\label{aggregation_special_case}
\begin{split}
    \mathbf{m}_{j} &= \textsc{aggregate}(\{\mathbf{h}_{n} \vert v_{n} \in\mathcal{N}(v_{j}) ) \\
    &= \sum_{v_{n}\in\mathcal{N}(v_{j})} \alpha_{jn} \mathbf{h}_{n},
\end{split}
\end{equation}
where $\alpha_{jn}$ is the trainable weight of edge $(n,j)$, and $j=1,2,\cdots,M$ indicates that the aggregation is performed among joints within the current pose rather than the whole skeleton sequence, which forms a spatial graph convolution as previously discussed. And because the trainable weights are shared by all samples, it is a static aggregation method that cannot learn to generate sample-specific messages that characterize each sample. Eq. (\ref{aggregation_special_case}) can be generalized to Eq. (\ref{aggregation}) in DD-GC with improved representational capacity.

\paragraph{Update Methods}

The aggregation step in GCN-based methods has generally received the most attention from researchers---in terms of proposing novel architectures and variations \cite{hamilton2020graph}. Next we turn our attention to the update step. The update operations in most of the existing approaches involve a projection matrix to project low dimensional representations to latent feature space \cite{hoff2002latent}, and the projection matrix is shared by all nodes in the graph. It is formulated as:
\begin{equation}\label{update_special_case}
\begin{split}
    \mathbf{h}_{j} & = \textsc{update}(\mathbf{h}_{j}, \mathbf{m}_{j}) \\
    & = \sigma\big(\mathbf{\Theta} (\mathbf{h}_{j} + \mathbf{m}_{j}) + \mathbf{b}\big),
\end{split}
\end{equation}
where $\mathbf{\Theta}\in\mathbb{R}^{d_{\text{in}} \times d_{\text{out}}}$ is the trainable projection matrix shared by all nodes. Eq. (\ref{update_special_case}) can be generalized to Eq. (\ref{update}) in DD-GC with joint-specific projection.

\section{Experiments}

    \begin{table*}[!ht]
        \centering
        \caption{MPJPEs of different actions for extended long-term prediction on H3.6M.}
        \label{long_pred}
        \resizebox{\textwidth}{!}{
        \begin{tabular}{c|cccc|cccc|cccc|cccc} \hline
        scenarios    & \multicolumn{4}{c|}{Walking} & \multicolumn{4}{c|}{Eating} &  \multicolumn{4}{c|}{Smoking} & \multicolumn{4}{c}{Discussion} \\ \hline
        milliseconds & 1400 & 1520 & 1680 & 2000    & 1400 & 1520 & 1680 & 2000   & 1400 & 1520 & 1680 & 2000     & 1400 & 1520 & 1680 & 2000       \\ \hline
        LTD	&44.06 	&49.47 	&60.54 	&65.41 	&86.32 	&80.56 	&79.79 	&91.09 	&77.61 	&80.79 	&84.29 	&89.32 	&98.08 	&104.05 	&106.27 	&116.74 \\
        MSR-GCN &47.14 	&50.98 	&60.38 	&64.56 &81.73 	&77.04 	&83.96 	&93.82 &86.84 	&91.97 	&95.82 	&100.80 &119.05 	&120.58 	&123.80 	&130.29 \\
        PGBIG	&46.30 	&51.66 	&58.91 	&68.63 	&86.76 	&83.31 	&83.70 	&100.16 	&77.76 	&83.40 	&89.24 	&92.40 	&105.71 	&106.92 	&111.44 	&116.76 \\
        BSTG-Trans	&51.42 	&56.21 	&65.19 	&75.36 	&87.55 	&89.54 	&90.32 	&99.09 	&78.53 	&84.36 	&91.17 	&94.79 	&111.93 	&112.98 	&114.98 	&115.78 	\\
        Ours	&56.53 	&59.15 	&65.09 	&70.56 	&88.72 	&84.99 	&85.61 	&94.23 	&\textbf{76.63} 	&\textbf{80.40} 	&86.36 	&91.65 	&109.35 	&112.26 	&112.53 	&\textbf{112.34} \\ \hline
        
        scenarios    & \multicolumn{4}{c|}{Directions} & \multicolumn{4}{c|}{Greeting} &  \multicolumn{4}{c|}{Phoning} & \multicolumn{4}{c}{Posing} \\ \hline
        milliseconds & 1400 & 1520 & 1680 & 2000    & 1400 & 1520 & 1680 & 2000   & 1400 & 1520 & 1680 & 2000     & 1400 & 1520 & 1680 & 2000       \\ \hline
        LTD	&95.68 	&96.67 	&104.08 	&125.72 	&112.44 	&133.57 	&154.02 	&165.50 	&152.24 	&165.46 	&184.48 	&219.69 	&276.25 	&277.93 	&265.74 	&305.99 \\
        MSR-GCN &92.28 	&99.04 	&114.26 	&129.86 &124.26 	&146.06 	&158.62 	&176.05 &144.99 	&158.06 	&176.80 	&199.21 &290.25 	&290.81 	&278.86 	&285.14 \\
        PGBIG	&101.53 	&103.90 	&110.63 	&133.84 	&109.71 	&130.75 	&150.70 	&164.23 	&151.18 	&162.98 	&180.77 	&211.26 	&271.66 	&273.52 	&261.85 	&299.68 \\ 
        BSTG-Trans	&95.31 	&100.60 	&106.95 	&130.10 	&111.35 	&133.06 	&151.82 	&168.95 	&156.75 	&167.83 	&181.57 	&218.08 	&273.15 	&276.91 	&264.61 	&311.85 	\\

        Ours	&\textbf{90.76} 	&\textbf{91.92} 	&\textbf{97.32} 	&\textbf{116.42} 	&\textbf{109.44} 	&\textbf{129.08} 	&\textbf{150.20} 	&166.54 	&155.00 	&166.80 	&182.66 	&215.46 	&\textbf{271.26} 	&\textbf{273.03} 	&267.29 	&314.11 \\ \hline
        
        scenarios    & \multicolumn{4}{c|}{Purchases} & \multicolumn{4}{c|}{Sitting} &  \multicolumn{4}{c|}{Sitting Down} & \multicolumn{4}{c}{Taking Photo} \\ \hline
        milliseconds & 1400 & 1520 & 1680 & 2000    & 1400 & 1520 & 1680 & 2000   & 1400 & 1520 & 1680 & 2000     & 1400 & 1520 & 1680 & 2000       \\ \hline
        LTD	&122.83 	&125.34 	&132.83 	&119.62 	&133.80 	&136.90 	&144.89 	&169.28 	&198.25 	&206.61 	&215.99 	&228.97 	&135.46 	&140.65 	&151.18 	&162.91 \\
        MSR-GCN &105.89 	&103.20 	&114.61 	&109.19     &150.18 	&154.29 	&166.91 	&190.67 &252.46 	&264.69 	&270.18 	&266.87&163.53 	&167.50 	&170.53 	&176.53 \\
        PGBIG	&134.81 	&135.43 	&142.91 	&129.51 	&129.14 	&133.43 	&143.51 	&169.90 	&166.84 	&168.93 	&177.82 	&214.35 	&121.99 	&127.08 	&134.61 	&148.56 \\
        BSTG-Trans	&142.12 	&143.81 	&150.51 	&135.14 	&129.81 	&135.76 	&140.96 	&163.92 	&163.68 	&159.93 	&171.92 	&207.79 	&120.99 	&127.30 	&127.61 	&144.23 	\\

        Ours	&144.24 	&145.29 	&154.22 	&140.15 	&\textbf{124.72} 	&\textbf{126.23} 	&\textbf{132.77} 	&\textbf{158.72} 	&\textbf{149.96} 	&\textbf{150.64} 	&\textbf{159.41} 	&\textbf{194.60} 	&\textbf{114.40} 	&\textbf{116.47} 	&\textbf{122.27} 	&\textbf{137.62} \\ \hline
        
        scenarios    & \multicolumn{4}{c|}{Waiting} & \multicolumn{4}{c|}{Walking Dog} &  \multicolumn{4}{c|}{Walking Together} & \multicolumn{4}{c}{Average} \\ \hline
        milliseconds & 1400 & 1520 & 1680 & 2000    & 1400 & 1520 & 1680 & 2000   & 1400 & 1520 & 1680 & 2000     & 1400 & 1520 & 1680 & 2000       \\ \hline
        LTD	&164.59 	&158.45 	&144.56 	&154.12 	&187.08 	&185.18 	&193.48 	&193.11 	&98.43 	&89.69 	&86.13 	&108.48 &132.21 &135.42 &140.55 &154.40\\
        MSR-GCN&152.65 	&147.24 	&144.71 	&161.55 &183.07 	&188.40 	&195.70 	&184.56 &95.99 	&89.83 	&93.65 	&116.71 &138.33 &141.96 &148.22 &159.05\\
        PGBIG	&163.70 	&156.48 	&145.77 	&159.24 	&182.57 	&181.23 	&182.88 	&170.83 	&93.28 	&87.98 	&84.78 	&109.53 &129.53 &132.47 &137.30 &152.59\\
        BSTG-Trans	&167.04 	&161.43 	&144.66 	&163.06 	&197.95 	&193.87 	&192.69 	&178.55 	&94.76 	&92.80 	&85.28 	&109.66 	&132.86 	&133.17 	&138.77 	&153.97 	\\

        Ours	&165.17 	&159.37 	&\textbf{144.00} 	&157.04 	&201.81 	&197.89 	&204.00 	&185.88 	&94.74 	&\textbf{87.06} 	&\textbf{83.47} 	&\textbf{101.35} &130.18 &\textbf{132.03} &\textbf{136.48} &\textbf{150.44}\\ \hline
        \end{tabular}}
    \end{table*}

    \begin{table}[t]
        \centering
        \caption{MPJPEs of four representative actions for normal long-term prediction on H3.6M.}
        \label{long_pred560_1000}
        \resizebox{\columnwidth}{!}{
        \begin{tabular}{c|cc|cc|cc|cc} \hline
        scenarios    & \multicolumn{2}{c|}{Directions} & \multicolumn{2}{c|}{Posing} &  \multicolumn{2}{c|}{Purchases} & \multicolumn{2}{c}{Sitting} \\ \hline
        milliseconds & 560 & 1000   & 560 & 1000& 560 & 1000& 560 & 1000    \\ \hline
        LTD	&79.51 	&103.34 	&109.88 	&208.63 	&92.09 	&122.40 	&84.57 	&115.64 \\ 
        MSR-GCN	&82.54 	&105.45 	&113.02 	&215.54 	&91.14 	&128.51 	&87.36 	&117.03 \\
        PGBIG	&89.28 	&110.73 	&104.37 	&205.63 	&94.17 	&125.72 	&78.28 	&111.92 \\ 
        BSTG-Trans & 85.10&108.92&118.36&209.77&92.34&126.18&82.02&115.25 \\
        Ours	&\textbf{74.31} 	&\textbf{100.23} 	&116.47 	&\textbf{196.90} 	&\textbf{90.72} 	&\textbf{121.15} 	&\textbf{76.33} 	&\textbf{109.17} 
        \\ \hline
        \end{tabular}}
    \end{table}

    \begin{table}[t]
        \centering
        \caption{MPJPEs of four representative actions for short-term prediction on H3.6M.}
        \label{short_pred320_400_h36m}
        \resizebox{\columnwidth}{!}{
        \begin{tabular}{c|cc|cc|cc|cc|cc} \hline
        scenarios    & \multicolumn{2}{c|}{Directions} & \multicolumn{2}{c|}{Greeting} &  \multicolumn{2}{c|}{Sitting} & \multicolumn{2}{c|}{Waiting} & \multicolumn{2}{c}{Average}\\ \hline
        milliseconds & 320 & 400   & 320 & 400& 320 & 400& 320 & 400& 320 & 400    \\ \hline
        LTD	&50.32 	&62.66 	&66.81 	&83.64 	&53.71 	&65.36 	&55.89 	&72.02 &56.68&70.92
        \\ 
        MSR-GCN	&52.65 	&62.07 	&66.86 	&87.20 	&51.46 	&62.30 	&57.65 	&74.20 &57.16&71.44
        \\
        PGBIG	&52.53 	&62.36 	&68.03 	&83.17 	&48.67 	&58.84 	&52.95 	&69.80 &55.54&68.54
        \\ 
        BSTG-Trans & 54.72&63.10&67.55&82.94&50.03&61.80&53.32&72.64&58.15&72.66 \\
        Ours	&\textbf{49.09} 	&\textbf{58.80} 	&66.98 	&\textbf{80.45} 	&\textbf{47.58} 	&\textbf{58.33} 	&\textbf{52.05} 	&\textbf{68.49} 
        &\textbf{53.93}&\textbf{66.52}
        \\ \hline
        \end{tabular}}
    \end{table}

    \begin{table}[t]
        \centering
        \caption{MPJPEs of three representative actions for both short- and long-term prediction on CMU Mocap.}
        \label{tab:short_pred80_1000_cmu}
        \resizebox{\columnwidth}{!}{
        \begin{tabular}{c|cc|cc|cc|cc} \hline
        scenarios    & \multicolumn{2}{c|}{Directing Traffic} & \multicolumn{2}{c|}{Running} &  \multicolumn{2}{c|}{Soccer} & \multicolumn{2}{c}{Average}\\ \hline
        milliseconds & 80 & 1000   & 80 & 1000& 80 & 1000& 80 & 1000\\ \hline
        LTD	&7.02 	&144.96 	&24.17 	&63.60 	&11.59 	&114.05 	&14.26 	&107.54    \\
        MSR-GCN &8.14   &149.68     &26.39  &64.35  &13.30  &117.64     &15.94  &110.56         \\
        PGBIG	&5.33 	&121.50 	&21.41 	&62.23 	&9.93 	&104.22 	&12.22 	&95.98     \\
        Ours	&\textbf{5.31} 	&\textbf{118.11} 	&22.16 	&\textbf{55.29} 	&\textbf{9.52} 	&107.56 	&12.33 	&\textbf{93.65} 
        \\ \hline
        \end{tabular}}
    \end{table}

    \begin{table}[t]
        \centering
        \caption{Comparison of efficiency in terms of inference delay, computational complexity, train time, and memory consumption.}
        \label{tab:efficiency}
        \resizebox{\columnwidth}{!}{
        \begin{tabular}
        {c|c|c|c|c} \hline
        Model   & Inference (ms) & FLOPs (M)  & Train (h) & Memory (GB)\\ \hline
        LTD      &27.23& 133.3     & 3     & 2.2  \\
        HisRep   &28.45&148.7   &5    &2.3\\
        MSR-GCN  &49.74& 1449.6    & 9     & 5.1 \\
        PGBIG    &30.26& 224.8     & 5     & 2.5 \\ \hline
        Ours     &29.50& 179.8     & 5     & 2.9 \\ 
        Ours (Single-level) &\textbf{25.81}& \textbf{121.3}     & \textbf{3}  & \textbf{2.2} \\  \hline
        \end{tabular}}
    \end{table}

Most current methods focus on anticipating the very recent future, making longer-term predictions over more than just a few frames is a more relevant and challenging task with many practical applications. Long-term prediction involves more uncertainties and lack of information, therefore posing greater challenges for the design of predictive models. To this end, we extend the concept of long-term beyond its conventional connotation in previous work \cite{tang2018long,cao2020long} to include longer time-scale, which highlights the need for more powerful models.

\subsection{Datasets}
(a) \textbf{Human3.6M (H3.6M)} \cite{ionescu2013human3} is a large-scale dataset which involves 15 types of actions performed by 7 actors. Following previous work, we preprocess the data to remove global rotation and translation, discard redundant joints, and downsample the sequence to 25fps. Then we convert the data to 3D coordinates. Each pose has 22 joints. We use subject 11 and subject 5 for validation and testing respectively, and the rest for training.
(b) \textbf{AMASS} \cite{mahmood2019amass} is a collection of multiple Mocap datasets unified by SMPL parameterization. We follow \cite{mao2020history,guo2023back} to use AMASS-BMLrub as the test set and split the rest of the AMASS dataset into training and validation sets.
(c) \textbf{3DPW} \cite{von2018recovering} is a dataset including indoor and outdoor scenes. A pose is represented by 26 joints, but we follow \cite{mao2020history,guo2023back} and evaluate 18 joints using the model trained on AMASS to evaluate generalization.
(d) \textbf{NTU RGB+D} \cite{shahroudy2016ntu} is contains 56,880 skeleton action sequences completed by one or two performers and categorized into 60 classes. It provides the 3D spatial coordinates of 25 joints for each human in an action.
(e) \textbf{CMU Motion Capture (CMU Mocap)} consists of 5 general action categories. Following previous work, we use 8 specific types of actions, and perform data preprocessing such that each pose consists of 25 joints.

    \begin{figure*}[t]
        \centering
        \includegraphics[width=\textwidth]{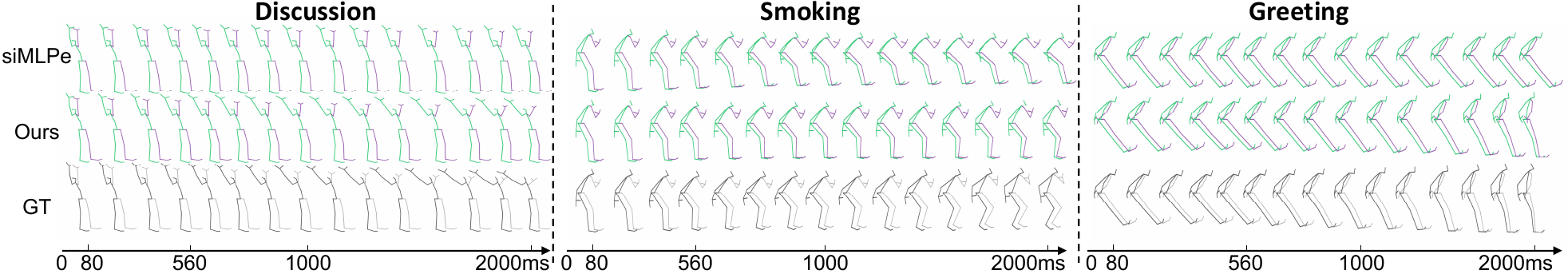} 
        \vspace{-1.5em}
        \caption{Qualitative comparisons through pose visualization on three representative actions: discussion, smoking, and greeting in H3.6M for extended long-term prediction. The colors green and purple respectively represent the left and right side of the human body. ``siMLPe'', ``Ours'', and ``GT'' respectively represent the motion sequences predicted by siMLPe~\cite{guo2023back} and our method, and their corresponding ground truths.}
        \label{pose_viz}
    \end{figure*}

    \begin{table}[!t]
        \centering
        \caption{Ablation on multi-level body structures.}
        \label{ablation_scale}
        \resizebox{\columnwidth}{!}{
        \begin{tabular}{cccccccc|ccc} \hline
        \multicolumn{8}{c|}{Joint Numbers} & \multicolumn{3}{c}{MPJPE (mm)} \\ \hline
        22&11&9&7&6&5&3&2 &80ms&1000ms&2000ms\\ \hline
        \multicolumn{11}{c}{Using 2 levels} \\ \hline
        \checkmark	&\checkmark	&	&	&	&	&	&	&9.69 	&110.38 	&151.74 \\
        \checkmark	&	&\checkmark	&	&	&	&	&	&9.74 	&110.42 	&151.55 \\
        \checkmark	&	&	&\checkmark	&	&	&	&	&9.76 	&110.57 	&151.60 \\
        \checkmark	&	&	&	&\checkmark	&	&	&	&9.92 	&110.78 	&151.93 \\
        \checkmark	&	&	&	&	&\checkmark	&	&	&9.80 	&110.77 	&152.47 \\
        \checkmark	&	&	&	&	&	&\checkmark	&	&9.97 	&111.03 	&152.85 \\
        \checkmark	&	&	&	&	&	&	&\checkmark	&10.03 	&111.15 	&151.99 \\ \hline
        \multicolumn{11}{c}{Using 3 levels} \\ \hline
        \checkmark	&\checkmark	&	&	&	&	&	&\checkmark	&\textbf{9.62} 	&\textbf{110.07} 	&\textbf{150.44} \\
        \checkmark	&\checkmark	&	&	&	&	&\checkmark	&	&9.84 	&110.11 	&150.58 \\
        \checkmark	&	&\checkmark	&	&	&	&	&\checkmark	&9.99 	&110.28 	&151.20 \\
        \checkmark	&	&	&\checkmark	&	&	&	&\checkmark	&9.89 	&110.71 	&151.49 \\
        \checkmark	&	&	&	&\checkmark	&	&	&\checkmark	&9.82 	&110.66 	&150.57 \\
        \checkmark	&	&	&	&	&\checkmark	&	&\checkmark	&9.77 	&110.54 	&152.10 \\ \hline
        \multicolumn{11}{c}{Using 4 levels} \\ \hline
        \checkmark	&\checkmark	&	&	&\checkmark	&	&	&\checkmark	&10.24 	&111.74 	&151.51 \\
        \checkmark	&	&\checkmark	&	&\checkmark	&	&	&\checkmark	&11.33 	&111.58 	&151.00 \\
        \checkmark	&\checkmark	&	&	&\checkmark	&	&\checkmark	&	&10.95 	&112.07 	&153.78 \\ \hline
        \end{tabular}}
    \end{table}

\subsection{Evaluation Settings}
Following the evaluation paradigm in previous work, we report the experimental results in terms of the Mean Per Joint Position Error in millimeter.
We compare our DD-GCN with ERD~\cite{fragkiadaki2015recurrent}, Res-RNN~\cite{martinez2017human}, Skel-TNet~\cite{guo2019human}, convSeq2Seq~\cite{li2018convolutional}, LTD~\cite{mao2019learning}, MSR-GCN~\cite{dang2021msr}, DMGNN~\cite{li2020dynamic}, HisRep~\cite{mao2020history}, PGBIG~\cite{ma2022progressively}, and siMLPe~\cite{guo2023back}.
Because they do not provide the results of extended long-term prediction, for fair comparison we re-train them separately to obtain the results for three different settings. We adopt the same testing strategy as previous state-of-the-art methods \cite{mao2019learning,li2020dynamic,martinez2017human}.
We implemented our DD-GCN using Pytorch, and we used Adam optimizer \cite{kingma2014adam} with leanring rate 0.00001 with a 0.96 decay every four epochs. Our models were trained on NVIDIA RTX 3080 Ti GPU for 50 epochs, the batch size was 16, and the gradients were clipped to a maximum $\ell_2$-norm of 1.

\subsection{Comparison with State-of-the-Art Methods}
Consistently with the discussion above, in evaluation of the proposed DD-GCN we report the experimental results for motion prediction in extended long-term, normal long-term and short term on H3.6M and CMU-Mocap datasets. In these cases we are given 10 frames (400ms) to predict the future 10 frames, 25 frames and 50 frames, respectively. We also evaluate our method on AMASS, 3DPW, and NTU RGB+D datasets in regular setting. We further illustrate the predicted samples for qualitative evaluation.

    \begin{table}[!t]
        \centering
        \caption{Ablation on numbers of DD-GC blocks.}
        \label{ablation_number_block}
        \resizebox{\columnwidth}{!}{
        \begin{tabular}{c|cc|cccc} \hline
         & \multicolumn{6}{c}{MPJPE (mm)} \\ \hline
        \# Blocks & 560ms &1000ms &1400ms &1520ms &1680ms &2000ms\\ \hline
        
        1	&80.48 	&112.54 	&132.51&133.99	&138.10&153.10 \\
        2	&\textbf{78.41} &110.52	&130.23&132.81	&136.99&150.58 \\
        3	&78.82 	&\textbf{110.07} 	&\textbf{130.18}&\textbf{132.03} 	&\textbf{136.48}&\textbf{150.44} \\
        4	&79.10 	&110.86 	&130.76&132.08	&136.15&150.96 \\
        5	&79.85 	&111.92 	&131.10&133.12	&137.59&152.49
        \\ \hline
        \end{tabular}}
    \end{table}

    \begin{table}[!t]
        \centering
        \caption{Ablation on dynamic message passing.}
        \label{ablation_cross_level}
        \resizebox{\columnwidth}{!}{
        \begin{tabular}{c|cccc} \hline
         & \multicolumn{4}{c}{MPJPE (mm)} \\ \hline
        Model Architecture &1400ms &1520ms &1680ms &2000ms\\ \hline
        DD-GCN &130.18&\textbf{132.03}&\textbf{136.48}&\textbf{150.44}\\
        DD-GCN w/o dynamic  &130.88&133.79 &137.15 &152.97\\
        DD-GCN w/o scaling 	&\textbf{129.33}&132.67&136.63&151.95\\
        DD-GCN w/o CLMP	&129.94 	&132.89 	&136.64 &152.83
        \\ \hline
        \end{tabular}}
    \end{table}

    \begin{table}[!t]
        \centering
        \caption{Ablation on 4D adjacency modeling.}
        \label{ablation_dense_4D}
        \resizebox{\columnwidth}{!}{
        \begin{tabular}{c|cccc} \hline
         & \multicolumn{4}{c}{MPJPE (mm)} \\ \hline
        Graph &1400ms &1520ms &1680ms &2000ms\\ \hline
        Dense Graph w 4D adjacency &130.18&132.03&\textbf{136.48}&\textbf{150.44}\\
        Separate Graph w 2D adjacency  &\textbf{129.90}&\textbf{132.01}&137.83&152.86
        \\ \hline
        \end{tabular}}
    \end{table}

\textbf{Long-Term Prediction on H3.6M.}
Normal long-term prediction is specifically defined as prediction over 500--1000ms, which may not be long enough to test the predictive power of a model to its full potential. Therefore we extend it to 2000ms and present the evaluation results for both normal and extended long-term prediction on H3.6M in Table \ref{avg_long_pred} and Table \ref{long_pred}. Table \ref{avg_long_pred} presents the results in terms of the more representative average prediction error for all actions at different times. It shows that DD-GCN in general achieves the best performance on average, especially as time grows, which testifies to its power regarding longer-term prediction. Table \ref{long_pred} presents the results in detail for different actions in extended long-term setting. We see that DD-GCN achieves better performance in cases of various difficult actions such as directions, greeting, sitting, sitting down, taking photo and walking together. Table \ref{long_pred560_1000} presents four representative actions in H3.6M for normal long-term prediction.

\textbf{Short-Term Prediction on H3.6M.}
Short-term prediction aims at prediction within 500ms. Following, we also report the short-term prediction on H3.6M in Table \ref{short_pred320_400_h36m}. Comparing DD-GCN to baselines on four representative actions, we see that DD-GCN achieves the best performance on average.

    \begin{figure}[t]
        \centering
        \includegraphics[width=\columnwidth]{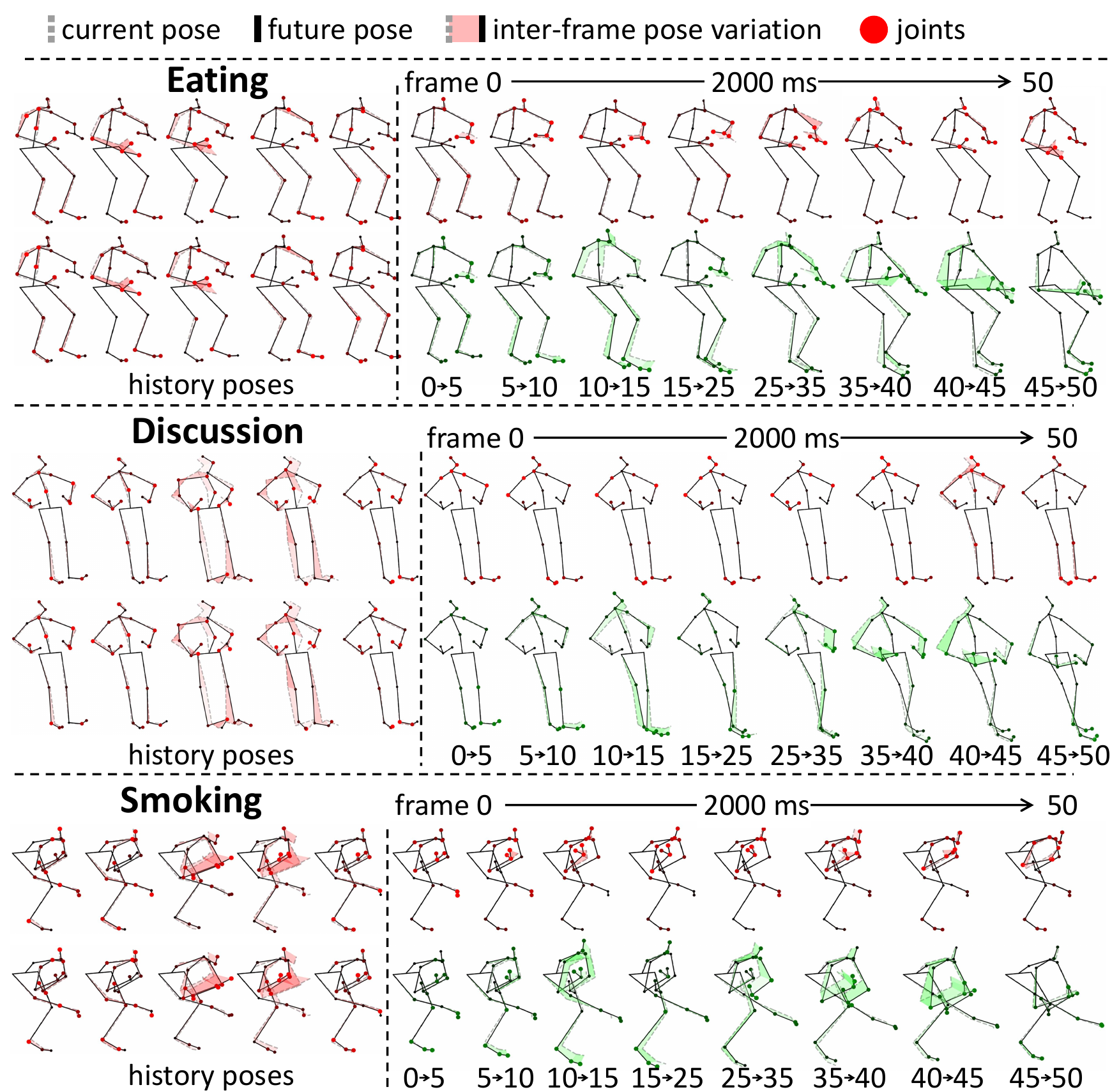}
        \caption{Failed cases of DD-GCN. The inter-frame pose variation is assessed by the size and color of points representing each joint, and by the area each bone sweeps through in 3D space between two frames. Points with lighter color and larger size indicate more intense joint movement.}
        \label{fig:failed cases}  
    \end{figure}
    
\textbf{Prediction on Other Datasets.}
Beyond H3.6M, we further consider the challenges of larger datasets or more complex real-world scenarios, and evaluate our method in these scenarios to test its predictive capability. Specifically, we further evaluate our method on AMASS, 3DPW, NTU RGB+D, and CMU Mocap datesets in regular settings. As existing works use different evaluation datasets and setups, and may not provide code for other datasets, we use different baselines in the tables, which is consistent with their original setup. 

We evaluate our method on AMASS and 3DPW following the testing paradigm as in \cite{guo2023back,mao2020history}. Experimental results in Tables \ref{tab:amass} and \ref{tab:3dpw} show that our model achieves state-of-the-art performance on AMASS and 3DPW datasets. We believe that testing our proposed method on these datasets brings valuable insights to the broader scope of our work. By extending our evaluation to include AMASS and 3DPW, we aim to provide a comprehensive assessment of our method's adaptability to diverse and challenging motion scenarios.
We also evaluate our model on CMU Mocap in Table \ref{tab:short_pred80_1000_cmu}.

We also evaluate our method on NTU RGB+D, a well recognized dataset for human action recognition and pose estimation. Although barely used in human motion prediction, we still evaluate our method on this dataset to test the generalization ability of our method. Considering the unique characteristics of NTU RGB+D dataset such as containing multi-person motion, we apply pre-processing techniques to the pose data in NTU RGB+D, including removing multi-person action and removing global translation. These steps are taken to ensure that the evaluation is conducted on a more comparable basis, providing a clearer understanding of our method's performance across different datasets.
Table \ref{tab:ntu} shows that our DD-GCN outperforms state-of-the-art methods by a large margin on NTU RGB+D dataset.

\textbf{Comparison of Efficiency.}
We evaluate the efficiency of our framework from different metrics, including inference delay, computational complexity, train time, and memory consumption. We compare the efficiency of our framework with state-of-the-art approaches in terms of these metrics. Table \ref{tab:efficiency} shows that the efficiency of DD-GCN is generally comparable to state-of-the-art approaches, while the single-level DD-GCN is the most efficient among all approaches.

\subsection{Ablation Study}

\textbf{Effects of Multi-level Body Structures.} To verify the proposed multi-level graph construction, we compare different settings in terms of the number of levels and joints. We design two additional scales which represent the skeleton using 9, 7, 6, 5, and 3 body parts respectively. Table \ref{ablation_scale} presents the MPJPEs result regarding our DD-GCN with different settings, which include 2, 3 and 4 levels. As shown in the table, our DD-GCN achieves the best performance on average with the original three levels with 22, 11 and 2 joints, especially in longer-term prediction.

    \begin{figure}[t]
        \centering
        \includegraphics[width=\columnwidth]{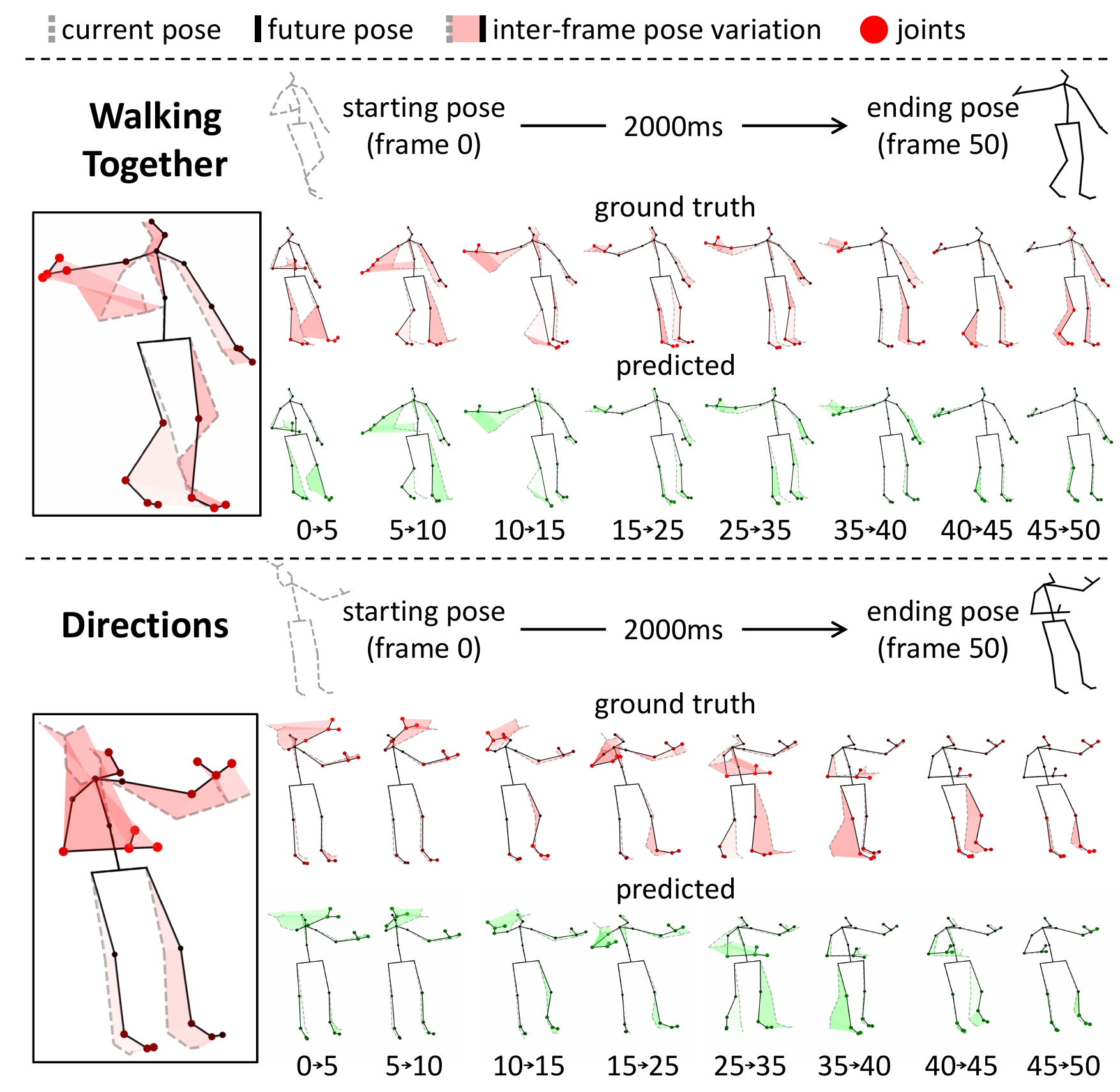}
        \caption{Successful cases of DD-GCN. The advantage of our DD-GCN lies in capturing distinctive patterns of different individuals.}
        \label{fig:successful cases}  
    \end{figure}

\textbf{Effects of Number of Blocks.}
To validate the effects of the dynamic dense graph convolution module, we modify the numbers of DD-GC blocks from 1 to 5 in our model and show the test results in Table \ref{ablation_number_block} for both normal long-term and extended long-term prediction. It shows that the prediction performance is the best when we employ 3 DD-GC blocks, while other numbers of DD-GCs show a decrease in performance, especially when the prediction time grows.

\textbf{Effects of Dynamic Message Passing.}
To validate dynamic message passing along with the cross-level feature fusion, we present the result of DD-GCN with dynamic feature aggregation removed, with cross-level message passing modules removed, and with the scaling operation removed, respectively. Note that in the case of DD-GCN w/o CLMP, we only preserve the cross-level message passing at the final layer before the final output. The results are shown in Table \ref{ablation_cross_level}.

\textbf{Effects of Dense Graph with 4D Adjacency Modeling.}
We study the effect of the dense graph with 4D adjacency modeling. We evaluate DD-GCN on separately defined trajectory graphs and pose graphs with 2D adjacency. It can be seen from Table \ref{ablation_dense_4D} that the proposed dense graph construction with 4D adjacency modeling results in the best performance, especially in extremely long-term prediction.

    \begin{figure*}[t]
        \centering
        \includegraphics[width=\textwidth]{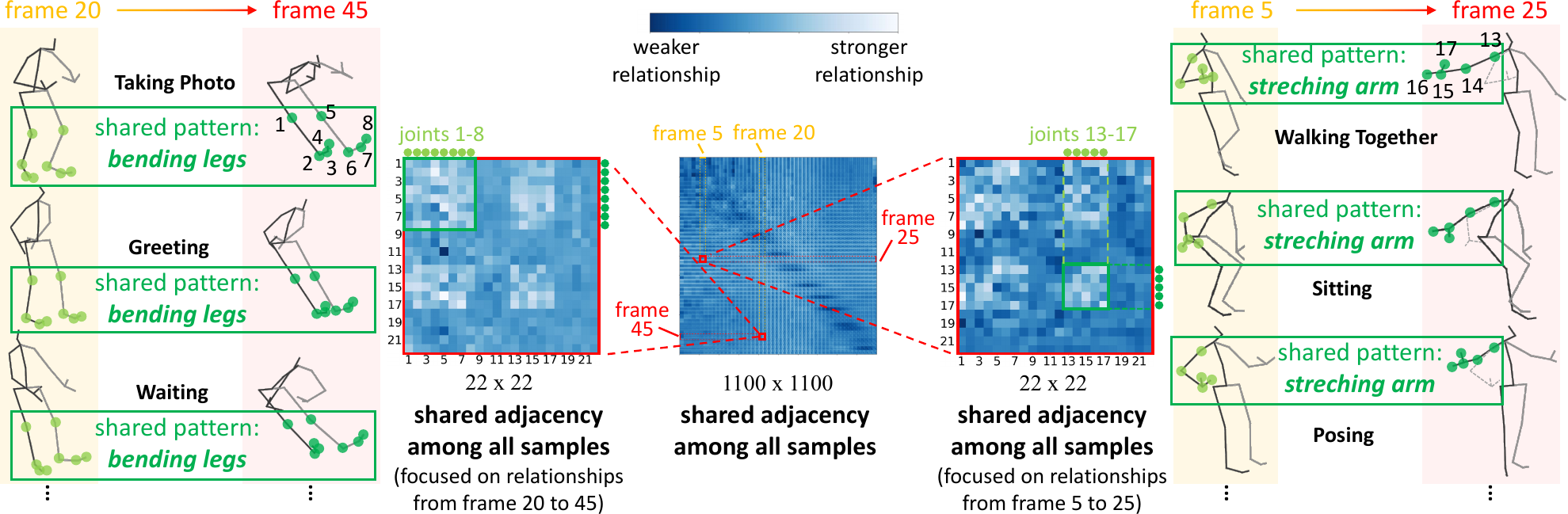}
        \caption{Visualization of shared adjacency and how it captures common patterns such as ``bending legs'' and ``stretching arm'' among various actions.}
        \label{fig:adj shared}  
    \end{figure*}
    \begin{figure}[t]
        \centering
        \includegraphics[width=\columnwidth]{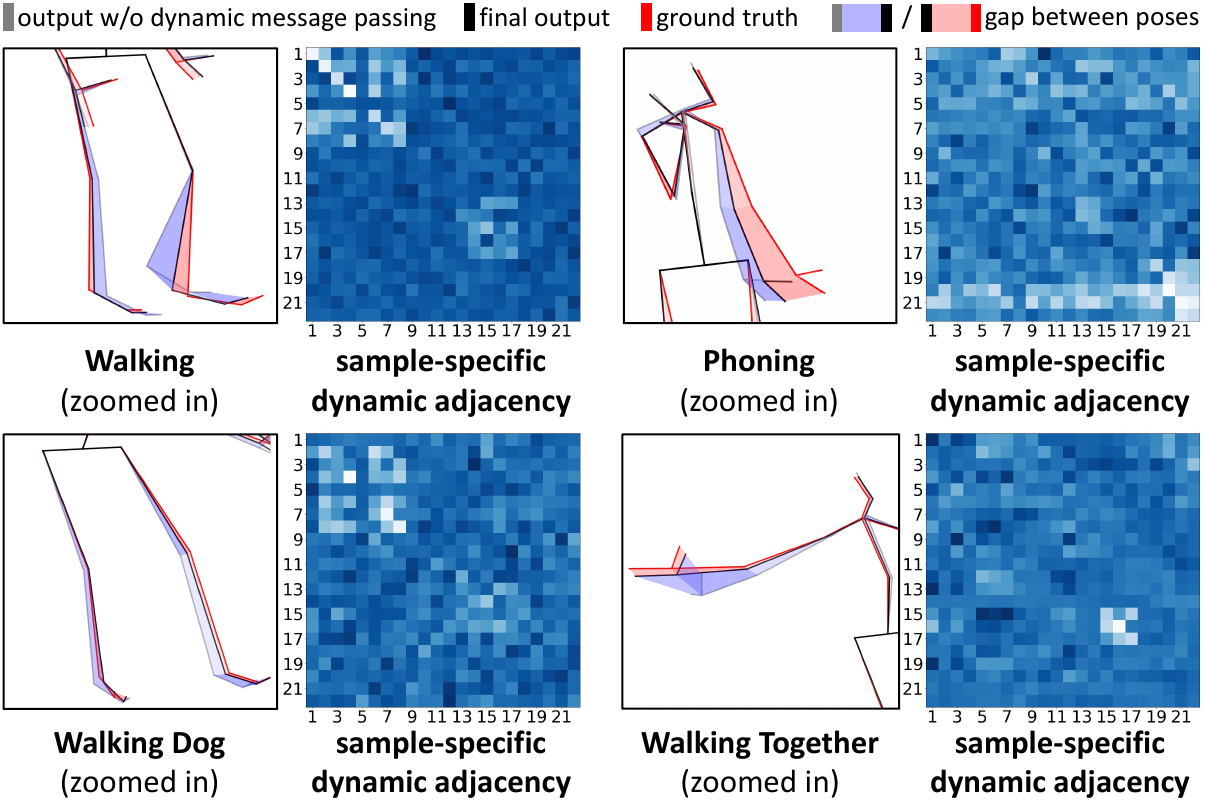}
        \caption{Visualization of the dynamic adjacency for different samples, and how it captures distinctive motion patterns in different action cases.}
        \label{fig:adj dynamic}  
    \end{figure}

\subsection{Visualization}

\textbf{Pose Visualization.}
In further comparison, we illustrate the future poses predicted by DD-GCN and the most recent state-of-the-art method, siMLPe~\cite{guo2023back}, on H3.6M. Figure \ref{pose_viz} provides the predicted poses of three actions: discussion, smoking, and greeting, throughout the whole time frame in extended long-term prediction. DD-GCN generates more realistic and accurate poses. The poses generated by siMLPe show imprecise body part movements, therefore accumulating large errors at the final pose. In general, DD-GCN yields better qualitative results, especially with the increase of prediction time.

\textbf{Qualitative Analysis.}
we have conducted comprehensive qualitative analysis, which shows that the advantage of our DD-GCN lies in capturing distinctive patterns of different individuals. As a result, our DD-GCN achieves notably superior performance on actions involving complex movements, and it yields limited performance on actions that are nearly still and thus do not exhibit distinctive patterns, which is uncommon in real-world scenarios. The qualitative analysis covers visualizations, case studies, and qualitative comparisons, to provide insights into the strengths as well as limitations of our method.

First, we visualize the failed cases to analyze the limitations of DD-GCN and aspects where it needs improvement. In Figure \ref{fig:failed cases}, we showcase three failed cases in which our model yields sub-optimal performance: eating, discussion, and smoking. Their pose sequences, along with the inter-frame pose variation, are visualized. The inter-frame pose variation is assessed by the size and color of points representing each joint, and by the area each bone sweeps through in 3D space between two frames. Points with lighter color and larger size indicate more intense joint movement. Areas that are less transparent indicate more intense bone movement. 
Figure \ref{fig:failed cases} shows that our DD-GCN yields sub-optimal performance in the cases where the history motion exhibits some kind of movements but the ground-truth future motion is nearly still. In these cases, our model tends to predict based on the observed history movements that the sample will continue to move in the future, rather than switch to a state of being still.

Next, we analyze the successful cases to gain a deeper understanding of our model's behavior and performance. To verify this, We show three successful cases respectively from eating, discussion, and smoking actions. The advantages of our model lie in that it can effectively capture distinctive patterns of each sample with dynamic message passing. Figure \ref{fig:successful cases} shows that our DD-GCN performs notably well on actions that are more complex and challenging, and involve visible movements such that they exhibit distinctive motion patterns, rather than actions composed of nearly-still poses.

\textbf{Adjacency Visualization.}
Figure \ref{fig:adj shared} demonstrates the research outcomes by visualizing the shared adjacency learned by the model, and how it captures common motion patterns among different samples. Lighter blue indicates stronger relationship between node pairs. The shared adjacency is originally a 4D array of size (50x22x50x22), where it stores the spatial-temporal inter-relationships among 50 frames across 22 joints. For better visibility, we reshape it to a matrix of size (1100x1100), and zoom in to show the details in different regions (sub-adjacency), which represent relationships between different frames. For example, the right side of the figure shows the sub-adjacency that stores the relationships from frame 5 to frame 25 across all joints. The shared adjacency is able to capture shared motion patterns among different samples such as ``stretching arm'', which contains large movements of joint 13 to joint 17. Therefore, joint 13--17 exhibit stronger relationships than other joints, which is consistent with the brighter area (row 13--17, column 13--17) in the sub-adjacency. Besides the pattern of ``stretching arm'', we show another pattern that the shared adjacency successfully captures: ``bending legs'' at the left side of Figure \ref{fig:adj shared}.

Figure \ref{fig:adj dynamic} demonstrates the research outcomes by visualizing the dynamic adjacency, and how it captures distinctive motion patterns of individual samples. To verify the necessity of dynamic adjacency, we first remove it from DD-GCN, i.e., remove the second sum term in Eq. (6), and then we test the DD-GCN respectively with and without dynamic message passing. Their differences from one another and from the ground-truth are highlighted by red and blue shaded area. The areas in blue represent the contribution of dynamic message passing to the final prediction in different action cases. We see that the dynamic message passing is able to adaptively adjust the poses based on each sample, driving the prediction notably closer to the ground truth.

\section{Conclusion}
This paper presents a global perspective for graph construction on human motion prediction task. Specifically, we propose a dense graph with 4D adjacency modeling to represent the motion sequence at different levels of abstraction as a whole. It enables the message passing to effectively capture complex spatio-temporal long-range dependencies, improving the expressive power of the network. Further, we propose a novel dynamic message passing framework, where we apply aggregators to generate informative messages by exploiting sample-specific relevance among nodes in the graph. The aggregator learns dynamically from data in a cost-effective way that provides higher interpretability and representational capacity at the expense of very few extra parameters. Overall, we propose a novel dynamic dense graph convolution (DD-GC) module that leverages both the two proposals. The DD-GC module is used as a building block for the overall network, termed DD-GCN, which adopts a multi-pathway framework with lateral connections that allow for direct and efficient feature learning at and across different levels. Extensive experiments verify that our DD-GCN outperforms state-of-the-art methods on both traditional long-term prediction task and our proposed extremely long-term prediction task.

\section{Acknowledgement}
This work was supported by the National Natural Science Foundation of China (No. 62203476).


\normalem
\bibliographystyle{IEEEtran}
\bibliography{references.bib}

\end{document}